\pdfoutput=1

\documentclass[11pt]{article}

\usepackage[final]{acl}

\usepackage{times}
\usepackage{latexsym}
\usepackage[T1]{fontenc}

\usepackage[utf8]{inputenc}

\usepackage{microtype}

\usepackage{inconsolata}

\usepackage{graphicx}

\usepackage{enumitem}
\usepackage{booktabs}
\usepackage{subcaption}
\usepackage{circuitikz}
\usepackage{pifont}
\newcommand{\cmark}{\ding{51}}%
\newcommand{\xmark}{\ding{55}}
\usepackage{amssymb}
\usepackage{amsmath}
\usepackage{bm}
\usepackage{xcolor}
\usepackage{multirow}
\usepackage{cleveref}
\usepackage{pdfpages}

\definecolor{yellow}{RGB}{230,159,000}
\definecolor{blue}{RGB}{0,114,178}
\definecolor{purple}{RGB}{046,037,133}
\definecolor{green}{RGB}{000,158,115}
\definecolor{pink}{RGB}{194,106,119}
\title{A Reality Check on Context Utilisation for \\ Retrieval-Augmented Generation}

\def\authorsep{\hspace{0.3em}}

\author{Lovisa Hagström\textsuperscript{{\includegraphics[height=1em]{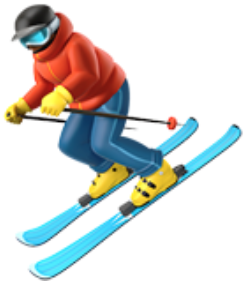}}} \authorsep Sara Vera Marjanović\textsuperscript{{\includegraphics[height=1em]{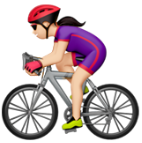}}} \\   
 \textbf{Haeun Yu\textsuperscript{{\includegraphics[height=1em]{figures/biker.png}}}} \authorsep \textbf{Arnav Arora\textsuperscript{{\includegraphics[height=1em]{figures/biker.png}}}} \authorsep \textbf{Christina Lioma\textsuperscript{{\includegraphics[height=1em]{figures/biker.png}}}} \\
 \textbf{Maria Maistro\textsuperscript{{\includegraphics[height=1em]{figures/biker.png}}}} \authorsep \textbf{Pepa Atanasova\textsuperscript{{\includegraphics[height=1em]{figures/biker.png}}}} \authorsep \textbf{Isabelle Augenstein\textsuperscript{{\includegraphics[height=1em]{figures/biker.png}}}} \medskip\\
\null\textsuperscript{\includegraphics[height=1em]{figures/skier.png}}Chalmers University of Technology \quad \null\textsuperscript{{\includegraphics[height=1em]{figures/biker.png}}}University of Copenhagen\\
\texttt{lovhag@chalmers.se}}

\begin{document}
\maketitle
\begin{abstract}
Retrieval-augmented generation (RAG) helps address the limitations of parametric knowledge embedded within a language model (LM). In real world settings, retrieved information can vary in complexity, yet most investigations of LM utilisation of context has been limited to synthetic text. 
We introduce \texttt{DRUID} (Dataset of Retrieved Unreliable, Insufficient and Difficult-to-understand contexts) with real-world queries and contexts manually annotated for stance. The dataset is based on the prototypical task of automated claim verification, for which automated retrieval of real-world evidence is crucial.
We compare \texttt{DRUID} to synthetic datasets (CounterFact, ConflictQA) and find that artificial datasets often fail to represent the complexity and diversity of realistically retrieved context.
We show that synthetic datasets exaggerate context characteristics rare in real retrieved data, which leads to inflated context utilisation results, as measured by our novel \textit{ACU} score. Moreover, while previous work has mainly focused on singleton context characteristics to explain context utilisation, correlations between singleton context properties and ACU on \texttt{DRUID} are surprisingly small compared to other properties related to context source.
Overall, our work underscores the need for real-world aligned context utilisation studies to represent and improve performance in real-world RAG settings.
\end{abstract}

\section{Introduction}

\begin{figure}[h]
    \centering
    \includegraphics[width=1.0\columnwidth,trim={1.9cm 3.9cm 25.3cm 0.3cm},clip]{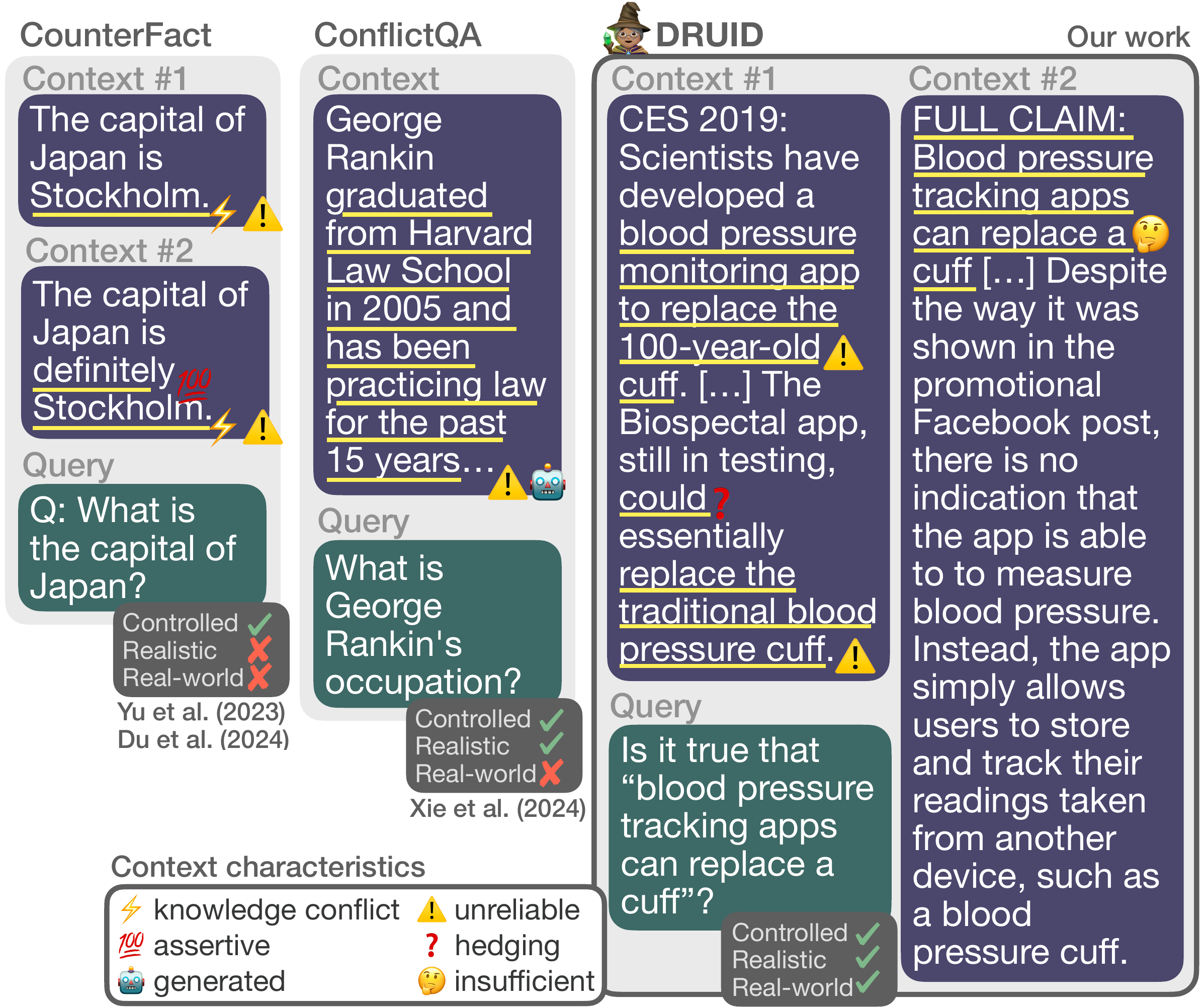}
    \caption{Datasets for context usage investigations.}
    \label{fig:comparison-overview}
\end{figure}

Retrieval-augmented generation (RAG) can be used to alleviate problems arising from imperfect parametric knowledge of language models (LMs), which may encode limited and potentially outdated information \citep{gao2024retrievalaugmentedgenerationlargelanguage,vu-etal-2024-freshllms}. However, the benefits of RAG are only realised if 1) the retrieval module retrieves helpful information and 2) the generative model successfully leverages the retrieved information. As a consequence, there have been many studies looking at the performance and interaction of these two components and how to improve them \citep{gao2024retrievalaugmentedgenerationlargelanguage}.

However, existing research has mainly studied RAG in a disjoint manner, where studies of the quality and relevance of the retrieved information are detached from studies of LM context usage \citep{shi2024irrelevant,xie2023adaptive,tan-etal-2024-blinded,du-etal-2024-context}. Hence, little is understood about 1) the characteristics of retrieved contexts and 2) their impact on LM context usage. Most notably, studies of LM context usage have leveraged controlled datasets using synthesised context to emulate a limited set of context characteristics (see \Cref{fig:comparison-overview}, left). For example, CounterFact and its variants are template-based
, lending a \emph{controlled} albeit very artificial and simplistic setup 
\citep{yu-etal-2023-characterizing,du-etal-2024-context}. ConflictQA, on the other hand, is based on a mix of generated and retrieved contexts to study context usage in a more \emph{realistic} setup with coherent and convincing contexts \citep{xie2024knowledgeconflict}. Nevertheless, the scenarios described by these datasets are not representative of \emph{real-world} RAG scenarios, as the context types do not reflect the diversity and complexity of the ones returned by an actual retriever present in RAG \citep{longpre-etal-2021-entity,ravaut-etal-2024-context,ortu-etal-2024-competition}. 

This work studies context usage for RAG in real-world scenarios with real-world queries and context, as opposed to artificial samples. 
To this end, we focus on the prototypical information-seeking task of fact verification, where retrieving and utilising real-world information is vital. 
For the task, an agent is provided with a statement about the world -- a \emph{claim} -- and needs to decide whether it is true or false using context retrieved from an external source -- \emph{evidence} \citep{guo-etal-2022-survey}. We take real fact-checked claims as `queries' and the retrieved evidence as `context' to evaluate RAG in this real-world setting, which naturally facilitates our goal of studying real-world context properties in RAG \citep{samarinas-etal-2021-improving,atanasova-etal-2022-fact,chrysidis2024credible,glockner2024ambifc}. 

In particular, this work makes three main contributions. First, we introduce \texttt{DRUID} (Dataset of Retrieved Unreliable, Insufficient and Difficult-to-understand context) with real-world (query, context) pairs to facilitate studies of context usage and failures in real-world scenarios (\S\ref{sec:dataset}). Second, we introduce a novel context-usage measure, \textit{ACU}, which rectifies issues in previous measures. Thirdly, we highlight major differences between popular synthetic datasets and real-world data (\texttt{DRUID}): both in over-arching characteristics (\S\ref{sec:characteristics}), as well as how the provided context is used across different popular LMs (\S\ref{sec:evaluation}). 

We show that synthetic datasets oversell the impact of certain context characteristics (e.g. knowledge conflicts), which are rare in retrieved data. Furthermore, synthetic data exaggerates the `context-repulsion' seen for LMs, as we rarely see this behaviour in realistic data. Finally, we show that there is no singleton context characteristic (e.g. context length or perplexity) indicating RAG failure in real-world settings. 
Altogether, our work provides a reality check on LM context usage and points to the need for real-world aligned studies to fully understand and improve context utilisation for RAG. We also provide tools and resources to facilitate such studies.\footnote{\url{https://github.com/copenlu/context-utilisation-for-RAG}}

\section{Related Work}

\paragraph{Claim Verification Datasets} Claim verification datasets typically measure an LM's ability to assess the veracity of a claim based on retrieved context (\emph{evidence}). 
Importantly, the information requirements of this task can be challenging, resulting in retrieved evidence that is noisy~\citep{samarinas-etal-2021-improving,atanasova-etal-2022-fact,fajcik-etal-2023-claim,chrysidis2024credible,glockner-etal-2022-missing,greta-chi}. Furthermore, as LMs fine-tuned for claim verification have been shown to ignore 
evidence~\citep{schuster-etal-2019-towards,schuster-etal-2021-get}, it is important to understand the causes of this behaviour. Therefore, with \texttt{DRUID}, we are the first to collect annotations for a range of `noisy' characteristics of retrieved real-world contexts to assess how they affect LMs. Furthermore, 
unlike concurrent claim verification datasets, which either present artificial samples or a limited or less realistic scenario for context retrieval \citep{thorne-etal-2018-fever,augenstein-etal-2019-multifc,diggelmann2020climatefever,averitec}, \texttt{DRUID} includes contexts \textit{automatically retrieved} from the web to assess their impact on RAG, leading to a wide diversity of context properties including \textit{insufficient} and \textit{leaked} evidence. No other existing fact-checking datasets fulfil all of these properties (see \Cref{tab:FCdatasets} in the Appendix).

\paragraph{Datasets for Context Usage Investigations}
Two popular datasets used for context usage investigations are CounterFact and ConflictQA \citep{meng2022locating,xie2024knowledgeconflict}. These datasets contain synthesised queries based on fact triplets from LAMA \citep{petroni-etal-2019-language} (e.g. Thomas Ong-citizen of-Singapore) for which some contexts have been synthesised to induce \emph{knowledge conflicts} by promoting answers in conflict with the parametric memory of the studied LM (e.g. `Pakistan' as opposed to `Singapore'). The datasets have found widespread use for work on mechanistic interpretability and the evaluation of context utilisation \citep{meng2022locating,geva-etal-2023-dissecting,ortu-etal-2024-competition}.

Similarly to CounterFact and ConflictQA, \texttt{DRUID} contains queries and corresponding contexts together with gold labels to facilitate evaluations of context utilisation. The main difference between \texttt{DRUID} and the two other datasets lies in how the queries and contexts were produced. For CounterFact and ConflictQA, the queries have been automatically synthesised based on WikiData subject-relation-object knowledge triplets (e.g. $\langle$George Larkin, occupation, lawyer$\rangle$ to produce ``George Larkin is a lawyer.'') and the contexts have either been synthesised based on an edited knowledge triplet or generated by an LLM prompted to produce alternative context supporting some edited knowledge triplet. This makes it easy to infer the gold labels for the synthesised contexts, while it is not representative of a real-world context usage scenario. DRUID, on the other hand, is based on queries sampled from naturally occurring claims and contexts from the web, retrieved by automated retrieval methods representative of a real-world RAG setup.

\paragraph{Impact of Context Characteristics for RAG}\label{sec:context-characteristics}

Work in information retrieval and RAG has identified several qualities in retrieved or synthesised contexts that impact context utilisation by humans and/or LMs. Retrievers typically provide overly long or corrupted text, which are \textit{difficult to understand}, and impact LM output \citep{gao2024retrievalaugmentedgenerationlargelanguage,vladika-matthes-2023-scientific}. Similarly, typos \citep{cho-etal-2024-typos} and high perplexity \citep{gonen-etal-2023-demystifying} have been identified as potential disruptors for RAG systems. Furthermore, \textit{implicit} contexts, lacking an explicit connection to the query, have been identified as a prevalent failure cause in RAG \citep{li-etal-2024-retrieval}. For automated retrieval situations, the rate of implicit contexts can be high due to chunking of text \citep{wang-etal-2024-dapr}. Instead, LMs have been shown to prefer context with high \textit{query-context similarities} \citep{wan-etal-2024-evidence}.

Most studies on RAG have focused on open-domain question answering \citep{kasai2023realtime,wu2024faithful}. \citet{yoran2023making,shi2024irrelevant} found that LMs are fragile to \textit{irrelevant information} in the context, harming performance. Furthermore, in the case of \textit{knowledge conflicts}, when context conflicts with parametric knowledge, LMs have been shown to ignore the conflicting context \citep{longpre-etal-2021-entity}, while other studies show that models prefer contextual information, as long as it is coherent and convincing \citep{xie2023adaptive}. \citet{sun2025explainingsourcesuncertaintyautomated} also connect knowledge conflicts to prediction uncertainty in fact-checking settings. Recently, \citet{xu-etal-2024-knowledge-conflicts} have proposed more granular categories for knowledge conflicts, using \emph{context-memory conflict} to denote the aforementioned phenomenon, and \emph{inter-context conflict} to refer to different contexts contradicting each other. \citet{marjanovic-etal-2024-dynamicqa} further study \textit{real-world knowledge conflicts} caused by dynamic facts, finding that RAG struggles the most with these.

\textit{Unreliable} contexts have been studied by \citet{chrysidis2024credible} in a fact-checking setup, for which misinformation is prevalent.
This type of information is typically overlooked in more generic RAG QA setups, potentially because the retrieval corpora usually are based on Wikipedia or pre-curated datasets.
\textit{References to external sources} may convince a human reader of the credibility of some context, yet LMs seem to be impervious \citep{wan-etal-2024-evidence}. However, expressed \textit{certainty/uncertainty} in text and its impact on LM context usage has recently been studied by \citet{du-etal-2024-context}, where assertive contexts are found to be more convincing.

In our creation of \texttt{DRUID} we combine all these aforementioned insights to annotate naturally occurring context characteristics of interest.

\section{DRUID} \label{sec:dataset}

Previous studies of context utilisation leverage synthetic datasets with synthesised claims and contexts, ignoring the retrieval part in RAG \citep{yu-etal-2023-characterizing,xie2024knowledgeconflict}. We develop the datasets \texttt{DRUID} (5,490 samples) and \texttt{DRUID+} (48,517 samples) to enable studies of context utilisation for real-world scenarios. To this end, we collect \textit{real-world claims from fact-checking sites} and use \textit{automated retrieval to fetch corresponding evidence from the web}. \texttt{DRUID} is a high-quality subset of \texttt{DRUID+} manually annotated for evidence relevance and stance. A \texttt{DRUID} sample consists of a $\langle$claim, evidence, labels$\rangle$ triple. More details on the dataset can be found in \Cref{tab:DRUID} and \Cref{app:DRUID}.

\begin{table}[t!]
    \centering
    \small
    \begin{tabular}{lrrrr}
    \toprule
    Source & \#claims & \#samples & IAA \\
    \midrule
    checkyourfact & 220 & 890 & 0.77 \\
    science.feedback & 220 & 913 & 0.64 \\
    factcheckni & 109 & 429 & 0.50 \\
    factly & 180 & 739 & 0.80 \\
    politifact & 220 & 931 & 0.74 \\
    srilanka.factcrescendo & 156 & 598 & 0.75 \\
    borderlines & 224 & 990 & 0.53 \\
    \midrule
    Total & 1,329 & 5,490 & 0.71 \\
    \bottomrule
    \end{tabular}
    \caption{Statistics for the \texttt{DRUID} dataset. IAA denotes inter-annotator agreement measured by Krippendorff's alpha. science.feedback also includes claims from climate.feedback and health.feedback.}
    \label{tab:DRUID}
\end{table}

\subsection{Claim Collection}

We sample claims verified by fact-checkers using Google's Factcheck API.\footnote{\url{https://developers.google.com/fact-check/tools/api/reference/rest}.} We only sample claims in English. The claims are collected from 7 diverse fact-checking sources, representing science, politics, Northern Ireland, Sri Lanka, the US, India, France, etc. All claims have been assessed by human fact-checkers. Further details on the claim collection can be found in \Cref{app:dataset}.

\subsection{Evidence Collection} 

For each claim in \texttt{DRUID} and \texttt{DRUID+}, we retrieve up to 5 and 40 snippets of evidence, respectively. First, a gold-standard evidence document is retrieved from the original fact-checking site, which is the `summary' of the fact-checking article written by the author of the article. For the remaining snippets of evidence, we use an automated retrieval method (\Cref{app:dataset}). We collect the top 20 search results for each of the Google and Bing search engines. The found webpages are then chunked into paragraphs and reranked by the Cohere rerank model.\footnote{\texttt{rerank-english-v3.0} from \url{https://docs.cohere.com/v2/docs/rerank-2}.} Evidence corresponding to the top-ranked chunks is included in \texttt{DRUID}.

\subsection{Relevance and Stance Annotation}

Since the evidence is collected using automated retrieval, as opposed to controlled synthesis, we need to assess the relevance of the retrieved information to the claim, and, if it is relevant, what stance it represents \citep{wang-etal-2024-factcheck}. For this, we crowd-source evidence-level annotations using Prolific\footnote{\url{https://www.prolific.com/}} and Potato \citep{pei-etal-2022-potato}. Each evidence piece in \texttt{DRUID} is double annotated for \textit{relevance} (\emph{relevant} or \emph{not relevant}) and \emph{stance} to the claim (\emph{supports}, \emph{insufficient-supports}, \emph{insufficient-neutral}, \emph{insufficient-contradictory}, \emph{insufficient-refutes} or \emph{refutes}). More details on the annotation, guidelines and examples from the annotation interface can be found in \Cref{app:annot-1}. 

The annotator compensation was approximately 9 GBP/hour (the compensation was fixed for each task while the annotator completion time varied).

\section{Context Characteristics}\label{sec:characteristics}

To understand the gap between the context provided in current diagnostic datasets for context usage and real RAG scenarios, we compare the characteristics present within our real-world dataset \texttt{DRUID} to the synthetic datasets CounterFact \cite{ortu-etal-2024-competition} and ConflictQA \cite{xie2024knowledgeconflict}. By virtue of their controlled setup, these and similar datasets have seen much use for the study of context utilisation and mechanisms thereof \citep{jin-etal-2024-cutting,du-etal-2024-context,tan-etal-2024-blinded,kortukov2024studying}.

To ensure adequate comparison, we recast all samples in CounterFact and ConflictQA to a claim-evidence format (see Appendix \ref{app:other_datasets}). This can be done without loss of information as all datasets represent a binary task for the LM (answer in alignment with the evidence or not). Furthermore, we show in \Cref{app:manipulation-results-after-reformat} that the analysis of context utilisation and mechanisms thereof are unaffected by the format of the task being either answer completion or claim verification -- the reformatting leads to no change in the mechanism employed by the model and its manipulation results.

In addition to the aforementioned datasets, we also present the characteristics of \texttt{DRUID+} to better understand the impact of only collecting the top-ranked evidence for \texttt{DRUID}.

\subsection{Detection of Context Characteristics}\label{sec:characteristics-detection}

Several context characteristics impacting context utilisation by humans and/or LMs have been identified by previous work (\Cref{sec:context-characteristics}). As opposed to synthesising contexts with certain properties, we \textit{detect} those in existing datasets. Along with manual annotation of relevance and stance, we leverage automated methods. We experiment with two types of automated detection methods to assess context characteristics: 1) rule-based methods and 2) prompt-an-LLM methods. For the latter we zero-shot prompt the Cohere Command R+ model.\footnote{\texttt{command-r-plus} (chat-only mode)} 

Initial trials leveraging human annotations of context characteristics showed high annotator disagreements, potentially due to the subjective nature of some of the characteristics, and were consequently abandoned. Instead, we opted to operationalise the properties, as we further describe below. This allows us to explore more model-based measures of context characteristics, which can be expected to have a greater impact on model context usage vis-a-vis subjective human perception of the same characteristics.

\paragraph{Relevance and stance} 
For \texttt{DRUID}, we use the manual relevance and stance annotations. For CounterFact and ConflictQA we infer those as follows. CounterFact contains counterfactual claims, for which the evidence is either the claim repeated (supports) or the claim but with the correct object restored (refutes).
For each ConflictQA entry, we have a model-generated claim and two types of evidence -- \emph{parametric memory aligned} (supports) or \emph{counter memory aligned} (refutes).  

\paragraph{Claim-evidence similarity} This is measured using Jaccard similarity (see \Cref{app:jaccard}), which outputs values between $[0,1]$, where 1 signifies maximum similarity. The overlap of claim words with evidence words, scaled by the number of claim words (`Claim-evidence overlap') is also measured. In addition, we detect if the evidence repeats the claim verbatim (`Repeats claim').

\paragraph{Difficult to understand} We measure the Flesch reading ease score, claim length (number of characters), evidence length (number of characters) and model context perplexities for our studied models (Llama 3.1 8B and Pythia 6.9B) to proxy how `difficult to understand' is the context. Generally, we may consider samples that correspond to high model perplexities to be confusing to the model.

\paragraph{Implicit} We detect named entities (NEs) in the claim and measure the overlap with entities found in the evidence (`Claim entity overlap'). spaCy \texttt{en\_core\_web\_trf} (based on RoBERTa-base) is used for the NE detection. Values are $\in [0,1]$ where 0 means that no NEs reappear in the evidence (maximum implicitness) and 1 means that all NEs were found in the evidence.

\paragraph{Refers to external source} Command R+ is prompted to tell whether some evidence contains a reference to an external source or not (`Detection by LLM'). Initial evaluation results show this detection method to align well with human annotations of the characteristic.

\paragraph{Uncertain} 
We use a lexicon-based approach proposed by \citet{islam-etal-2020-lexicon} to detect hedge words and hedging discourse markers in the evidence to proxy `uncertain' (`Contains hedging' and `Contains hedging discourse'). If a hedge word or hedging discourse marker is detected in the evidence, it is marked as `uncertain' according to that method. 

\paragraph{Unreliable} We use manually curated lists by Media Bias/Fact Check\footnote{\url{https://mediabiasfactcheck.com/}} (MBFC) to automatically detect whether the evidence piece originates from a web page marked as using questionable sources, promoting conspiracy/pseudoscience or being a satire site (`Unreliable source'). However, due to the sparsity of the MBFC lists, we are unable to detect unreliability for all evidence in \texttt{DRUID} and \texttt{DRUID+}, lacking results for 26\% and 34\% of the samples, respectively. For CounterFact and ConflictQA there are no evidence sources to analyse.

\paragraph{Additional characteristics}
We check whether the evidence can be seen as directly pointing out a verdict by measuring whether the evidence contains the word `True' or `False'. For the \texttt{DRUID} and \texttt{DRUID+} datasets, we also record whether the evidence was published after the claim was made, as this allows to measure the occurrence of and effects of leaked information (`Pub after claim') \citep{averitec}. Similarly, we measure whether the evidence comes from a fact-check webpage as this can be expected to contain additional leaked information (`Fact-check source') and whether it comes from the original fact-checking site summary (`Gold source').

\begin{figure}[t!]
    \centering
    \includegraphics[width=\linewidth,trim={7.8cm 5cm 36.2cm 2.5cm},clip]{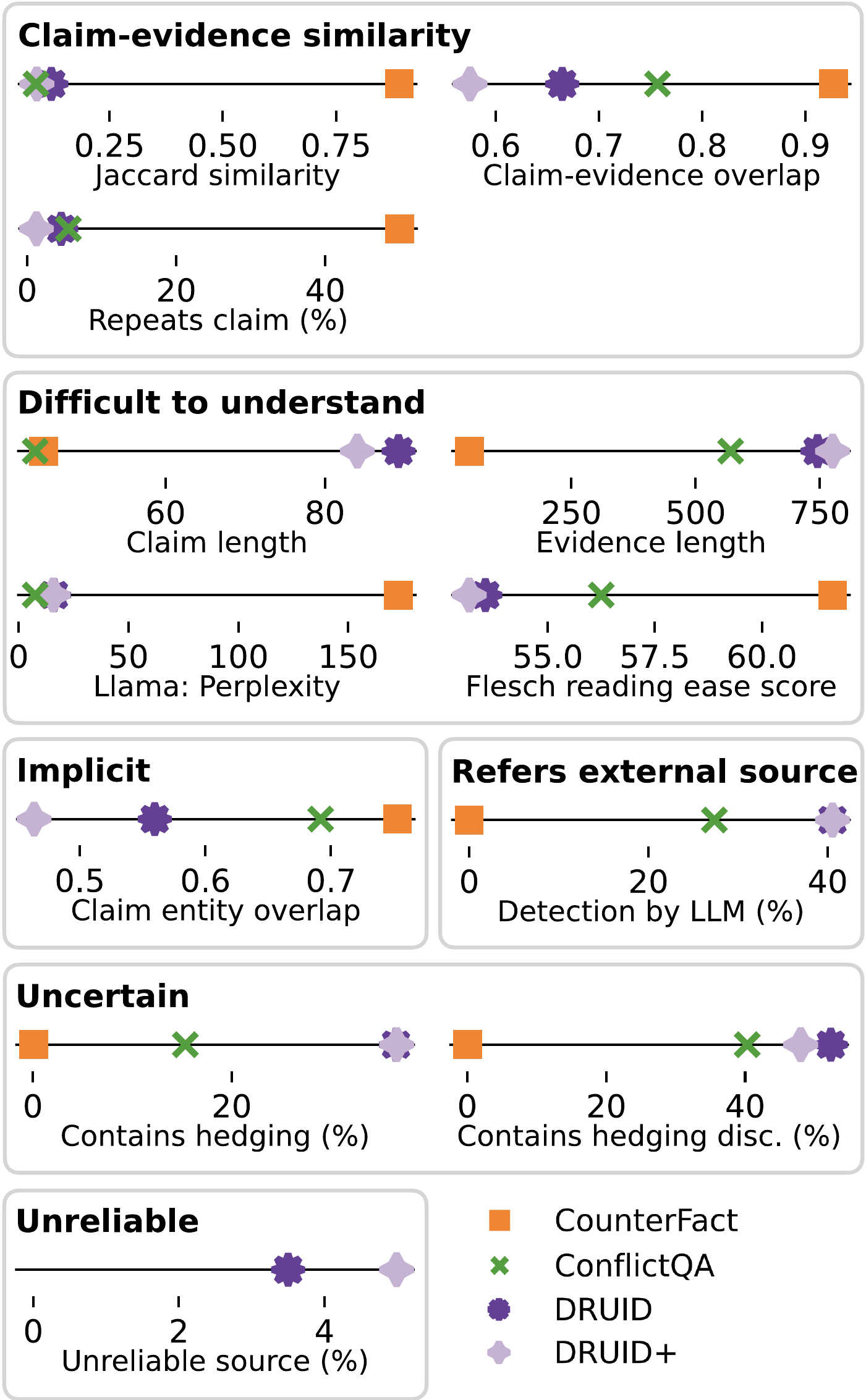}
    \caption{Average values for the context characteristics in CounterFact \citep{yu-etal-2023-characterizing}, ConflictQA \citep{xie2024knowledgeconflict} and \texttt{DRUID} datasets. The characteristics and their detection are described in \Cref{sec:context-characteristics,sec:characteristics-detection}, respectively.}
    \label{fig:characteristics-plot}
\end{figure}

\subsection{Analysis of Context Characteristics}

\paragraph{Relevance and stance}
Relevance and stance annotations for all datasets are shown in \Cref{tab:property-statistics-relevant,tab:property-statistics-stance} in the Appendix. Most contexts are annotated as relevant; however, given the more ambiguous nature of real-world queries, especially in claim verification, there is more variety in the kinds of stances presented by the context provided in \texttt{DRUID}: the majority of the automatically retrieved contexts (50\%) do not have a clear stance or are not sufficient for addressing the query. This is the consequence of using automated retrieval, for which not even state-of-the-art methods based on commercial search engines and Cohere modules are capable of consistently retrieving `gold standard context'. Admittedly, the retrieval setup is used in a zero-shot fashion and performance may improve somewhat with additional fine-tuning, while it would not solve all insufficiency issues stemming from automated retrieval. Conversely, synthesised samples always assume sufficient context. Our results show a clear discrepancy between synthesised and real-world datasets, proving the need for real-world aligned datasets for studies of context usage. 

\paragraph{\texttt{DRUID} in comparison with other RAG datasets} The detected context characteristics for the synthetic datasets and \texttt{DRUID} are shown in \Cref{fig:characteristics-plot}. More detailed results can be found in \Cref{app:characteristics}. The synthetic CounterFact dataset has the highest Flesch reading ease scores, significantly shorter evidence lengths, frequent repetitions of the claim in the evidence, significantly higher Jaccard similarity values and very few uncertainty markers relative to the other investigated datasets. CounterFact was designed to showcase simple knowledge conflict scenarios, causing much greater perplexity for both investigated LMs. In general, CounterFact is tailored to a specific type of context usage that is not indicative of real, retrieved context, as shown by the comparison to \texttt{DRUID}. The generated evidence for ConflictQA has characteristics more similar to those of \texttt{DRUID}.  However, we can see that \texttt{DRUID} has much longer claims and evidence than seen in either of the other datasets; furthermore, there are more uncertainty markers and a greater degree of implicitness in the naturally occurring context in the \texttt{DRUID} dataset.

\paragraph{Memory conflicts are less prevalent in real-world scenarios.} We measure context-memory conflicts by comparing the parametric model prediction (no context or evidence provided) with the stance of the provided evidence. We have a conflict if the prediction and stance are either `Refutes' or `Supports' and do not align. For Llama 3.1 8B we record memory conflicts on 97.41\% of the supporting evidence from CounterFact and on 71.16\% of the refuting evidence from ConflictQA. For \texttt{DRUID}, we identify memory conflicts for 58.09\% of the supporting evidence. Evidently, the rate of memory conflict is lower in real-world scenarios compared to artificial scenarios. More detailed results and the results for Pythia can be found in \Cref{tab:pred-results}.

\section{Context Utilisation} \label{sec:evaluation}

 We aim to assess the transferability of insights based on synthesised scenarios to real-world scenarios. To this end, we evaluate and compare LM context utilisation results on synthetic datasets to results on \texttt{DRUID}.

\subsection{Method}

We measure the context utilisation of Pythia 6.9B and Llama 3.1 8B, two models from two model families widely used in RAG-evaluation studies \cite{pmlr-v202-biderman23a,grattafiori2024llama3herdmodels,ortu-etal-2024-competition,xie2024knowledgeconflict,jin-etal-2024-cutting}, on the CounterFact, ConflictQA and \texttt{DRUID} datasets. To measure context utilisation, the models are evaluated in two modes: 1) without evidence and 2) with evidence. In both modes, the models are prompted to assess the veracity of a given claim (True, False, or None), without and with evidence respectively. More details on the prompting can be found in \Cref{app:prompts}. We evaluate context utilisation using the softmaxed model logits, which we describe further in the next section. In the main paper, we only show results for supporting and refuting evidence; behaviour for all forms of `insufficient' evidence (where `None' is the expected model output) can be found in \Cref{app:additional-cu-results,app:full-corr-results}.

\subsection{Evaluation} 

There is no consistent measure for context usage across similar work; many studies look simply at changes in overall output distributions \citep{du-etal-2024-context,marjanovic-etal-2024-dynamicqa}, which does not guarantee that the change is relevant to the provided context. Works in mechanistic interpretability often rely on logit differences for a specific token given evidence \cite{ortu-etal-2024-competition,yu-etal-2023-characterizing}, which are not normalised, do not factor in desired change, and limit comparisons. Due to these issues, we introduce a novel measure (\textit{ACU}), which 1) uses softmax-normalised probabilities, to ensure meaningful comparison, 2) focuses on probabilities of specific tokens, to ensure relevant change, and 3) scales these values by the amount of possible increase in probability. To measure context usage for a model $M$, we consider the re-scaled difference in salient token probability $t \in T = \{\mathrm{True}, \mathrm{None}, \mathrm{False}\}$ for a claim $C$ between settings with and without evidence $E$, as follows.
\begin{equation}
    \Delta P_M(t|C,E) = 
\begin{cases}
    \frac{P_M(t|C,E)-P_M(t|C)}{1-P_M(t|C)} \\ 
    \text{\small if } \scriptstyle P_M(t|C,E)\geq P_M(t|C), \vspace{0.3cm}\\
    \frac{P_M(t|C,E)-P_M(t|C)}{P_M(t|C)} \\ 
    \text{\small otherwise.}
\end{cases}
\label{eq:delta-p}
\end{equation}
Here, $P_M(t|C)$ and $P_M(t|C, E)$ denote the output probabilities for token $t\in T$ by model $M$ given a claim $C$ and evidence $E$, respectively. The rescaling ensures that our metric is less sensitive to the original $P(t|C)$ value.
We expect high positive values of $\Delta P_M(t|C,E)$ for $t$ that align with the stance of $E$ and the opposite for $t$ that conflict with the stance. For example, given an evidence piece with the stance \emph{refutes} we should ideally measure a high value for $\Delta P_M(\mathrm{False}|C,E)$ and low values for $\Delta P_M(\mathrm{True}|C,E)$ and $\Delta P_M(\mathrm{None}|C,E)$.

We define a score of accumulated context usage (ACU) per sample $\{C, E\}$ with stance $S_E$ for a model $M$ as follows.
\begin{align}\label{eq:acu}
    & \mathrm{ACU}(C, E, S_E, M) = &&\\\nonumber
    & = \frac{1}{|T|}\sum_{t\in T} D(t, S_E) \Delta P_M (t|C,E) &&
\end{align}
$D(t, S_E)$ denotes the desirable change in $\Delta P_M$ for maximum context usage, which is either \{-1,1\}, depending on the annotated stance of the evidence. For example, $D(\mathrm{False}, \mathrm{refutes}) = 1$, whereas $D(\mathrm{True},\mathrm{refutes})=D(\mathrm{None},\mathrm{refutes})=-1$. This limits the range of ACU between $[-1,1]$.

\begin{figure}[h]
    \centering
    \includegraphics[width=0.95\linewidth]{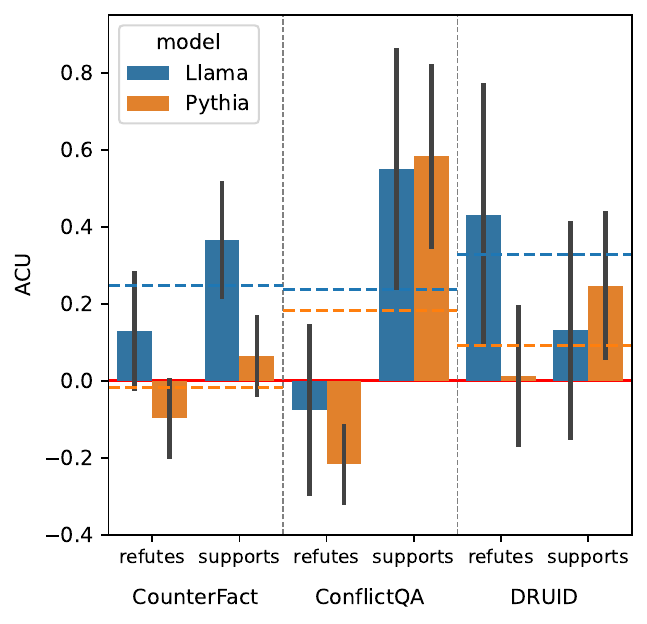}
    \caption{ACU (\Cref{eq:acu}) for each model and dataset. The error bars indicate the standard deviation. 
    Negative ACU values indicate `context-repulsion': changes in probability away from the annotated evidence stance.
    The dashed horizontal lines indicate average ACU scores for each model and dataset.}
    \label{fig:cu-agg}
\end{figure}

\subsection{How do LMs utilise real-world retrieved context compared to synthesised context?}

We inspect the context usage behavior of Pythia and Llama on CounterFact, ConflictQA and \texttt{DRUID} to understand how LMs utilise real-world context compared to synthetic contexts. Accumulated context usage scores (\Cref{eq:acu}) can be found in \Cref{fig:cu-agg}. See \Cref{app:additional-cu-results} for more granular context usage results. We structure the analysis around a set of main findings, listed below.

\paragraph{Synthetic datasets suggest an over-preference of supporting evidence.} While we see variations in model behaviour for our two synthetic datasets, ConflictQA and CounterFact, there are also some over-arching similarities: context utilisation is much greater for supporting evidence. In the case of refuting evidence, we often see negative ACU scores, indicating `context-repulsion', changes in probability \emph{away} from the stance of the provided context, which indicates low robustness; this is strongest for ConflictQA, which also has the greatest ACU scores for supporting context. This may be the consequence of the ConflictQA claims having been generated by Llama 2 7B and some of the supporting contexts having been generated by ChatGPT (they are \emph{aligned with parametric memory}, which has been shown to increase context utilisation \citep{xie2024knowledgeconflict,tan-etal-2024-blinded}). However, this preference for supporting evidence is also seen in CounterFact, which is surprising as the refuting evidence should align with LM parametric memory. This may be explained by the synthetic and confusing nature of CounterFact samples, leading to high model perplexities (See \Cref{fig:characteristics-plot}.)

We see a different behaviour with our real-world dataset \texttt{DRUID}: we rarely see context repulsion, and we see reduced ACU scores for supporting evidence. These lower ACU scores for supporting evidence may stem from the lack of generated context (vis-a-vis ConflictQA), and the increased ACU for refuting evidence may be due to the lower perplexities of the context (vis-a-vis CounterFact). 
This highlights the need for real-world contexts for studies of context utilisation: automatically generated contexts, by automated template-filling or LLM generation, inevitably induce properties that interfere with studies of context utilisation. 

\paragraph{Different models show different context usage.} Notably, Llama and Pythia behave very differently on all datasets studied. Potentially, this owes to CounterFact having been designed to elicit knowledge conflicts in Pythia and ConflictQA having been based on claims generated by Llama 2 7B. However, \texttt{DRUID} has not been customised to a specific model and results on this dataset clearly show how context usage varies across models. Moreover, we find that Llama on average is more faithful to the contexts of all datasets (and demonstrates less context-repulsion), yet remains understudied in context-utilisation studies \cite{ortu-etal-2024-competition, du-etal-2024-context}. These results are further corroborated in \citet{hagstrom-cub}, for which multiple LMs of different model family and size are benchmarked for their context utilisation.

\subsection{Does LM context usage depend on characteristics of the evidence/context?}

\begin{figure}
    \centering
    \includegraphics[width=\linewidth]{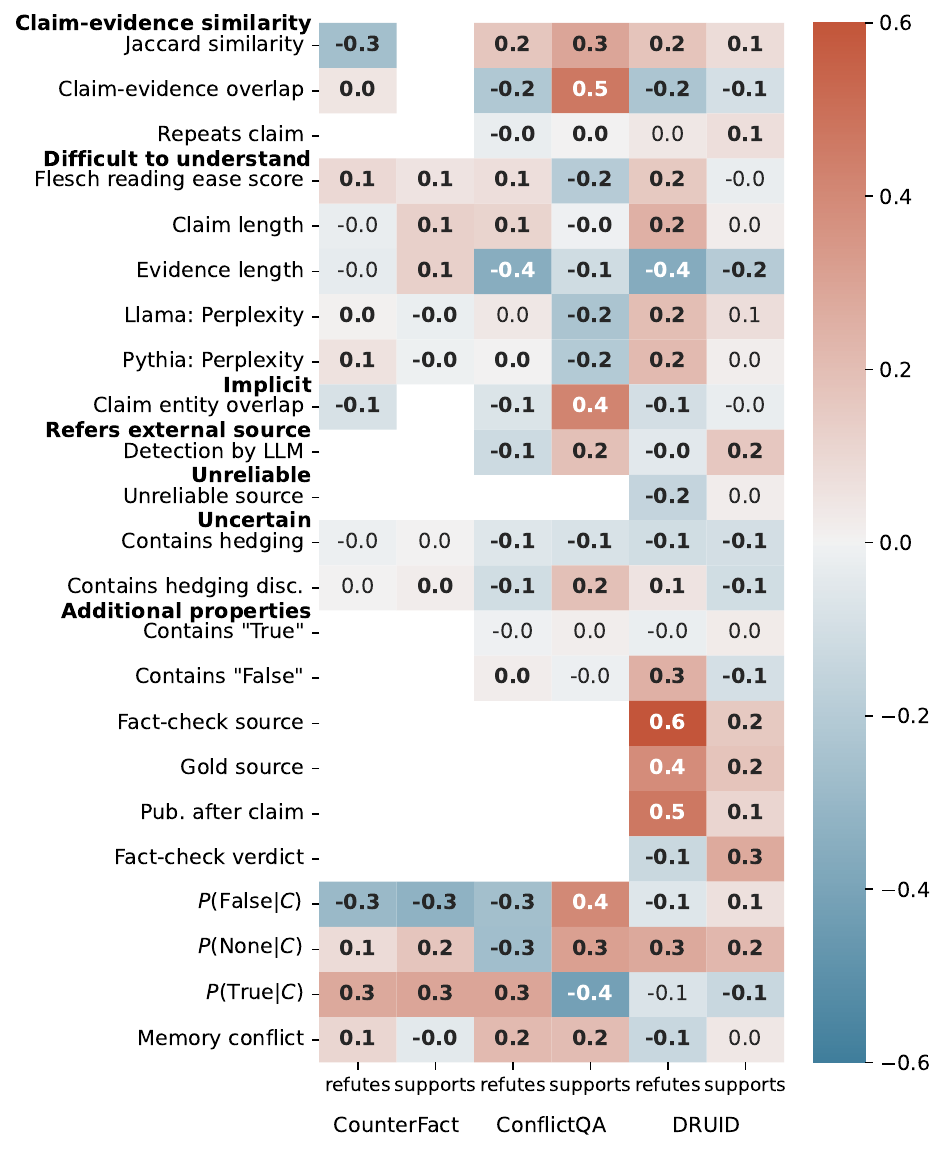}
    \caption{Spearman correlations between context usage measured by ACU (\Cref{eq:acu}) and different context characteristics for Llama. Significant correlation values (p-value < 0.05) are marked in \textbf{bold}.}
    \label{fig:corrs-Llama}
\end{figure}

We evaluate the influence of different context characteristics (see \S\ref{sec:characteristics-detection}) on model context usage. For this, we calculate Spearman correlations between each context property and our context usage metric, ACU (\Cref{eq:acu}), stratified by the evidence stance for each dataset. The results for Llama are shown in \Cref{fig:corrs-Llama}. Results for Pythia, insufficient evidence from \texttt{DRUID} and additional fine-grained correlation results can be found in \Cref{app:full-corr-results}. While we see a limited effect of any one characteristic, we highlight overarching findings below.

\paragraph{Context from fact-check sources have greater ACU scores.} Llama and Pythia are more likely to be faithful to refuting context from a fact-checking source. Most likely, this general property captures a characteristic not fully captured by the more fine-grained detection methods. Previous manual inspection of fact-check articles indicates a higher rate of assertive and to-the-point language, which may explain these observations. Moreover, the fact-check articles are more likely to directly discuss the claims in a one-hop manner with multiple arguments, possibly making them more convincing to the LM. We hypothesise that a similar case holds for `Pub. after claim' and `Gold source'. Similar results are observed for Pythia. The restricted generation process of the synthetic datasets makes similar investigations impossible. 

Previous studies of context utilisation have focused on the effects of singular context characteristics in isolation. Our work hints at the relevance of including aggregates of features in the analysis, as these may better explain context utilisation in real-world scenarios. For example, contexts from a certain source, like fact-checking articles, may express aggregates of features.

\paragraph{References to external sources show low correlations with ACU.} We measure low correlations with whether the evidence refers to an external source on both ConflictQA and \texttt{DRUID}.  Our results on ConflictQA and \texttt{DRUID} show slightly greater importance of references for supporting contexts, while the values measured are fairly low. Using a synthetic dataset, \citet{wan-etal-2024-evidence} found LM context usage to be insensitive to references to external sources. Our results on the real-world \texttt{DRUID} dataset support this conclusion.

\paragraph{Correlations with claim-evidence similarity properties are low for \texttt{DRUID}.} For ConflictQA we measure the highest correlation with context usage for claim-evidence overlap for supporting contexts. The same does not hold for \texttt{DRUID}.  Previous work by \citet{tan-etal-2024-blinded} on controlled context-conflicting datasets found LMs to prioritise contexts with high similarities between query and context. Our results indicate that real-world queries and contexts come with a greater complexity for which context usage cannot be predicted solely based on query-context similarity.

\paragraph{LMs are less faithful to long contexts.} Llama is less likely to be faithful to long refuting contexts on both \texttt{DRUID} and ConflictQA. We note how the results do not generalise to CounterFact, which might be explained by the more synthetic nature of CounterFact compared to ConflictQA.

\section{Conclusion}
In this work, we ground studies of context utilisation to real-world RAG scenarios. We develop \texttt{DRUID} and compare it to synthesised datasets previously used to study context-utilisation. \texttt{DRUID} is a claim-verification dataset which contains naturally occurring claims and manually annotated evidence automatically retrieved from the web.
We find fundamental differences in dataset characteristics between \texttt{DRUID} and synthetic datasets (CounterFact and ConflictQA). 
We also introduce a novel ACU score to consistently measure context utilisation across LMs and datasets.
On \texttt{DRUID}, correlations between singleton context properties and ACU are surprisingly small compared to other properties related to context source (e.g. contexts coming from specific types of websites). We hypothesise that, rather than singleton features, this owes to an aggregation of several characteristics contributing to context usage. This suggests the common factors impacting RAG success are broader than previously expected, and further work needs to be done to identify fine-grained causes of RAG failure. Furthermore, given the use of synthetic datasets to identify mechanistic components of context usage \cite{ortu-etal-2024-competition, yu-etal-2023-characterizing}, our results call into question the generalisability of the findings. With \texttt{DRUID}, we provide resources that better facilitate mechanistic and behavioural studies of context usage in real-world scenarios.

\section{Limitations}

Our work leverages claim verification as a vehicle for studies of realistic context utilisation. It is not fully clear whether insights related to context utilisation on this task will transfer to other RAG tasks, such as question-answering, which is overly represented in RAG evaluations. However, claim verification is a complex information-seeking task and we expect other tasks to have a large overlap or subset of properties with it. For example, as seen in this work, the QA format for CounterFact and ConflictQA is easily recast as a claim verification task. Furthermore, we show that intervention methods developed for QA tasks easily transfer to the same datasets when recast to a claim-verification setting in \Cref{app:manipulation-results-after-reformat}. This suggests that some findings can be generalisable across tasks. Future work could expand on this work to \textit{incorporate other RAG-specific tasks} to better understand the generalisability of context utilisation behaviours.

In our creation of \texttt{DRUID} we leverage an automated retrieval method based on commercial search engines and the Cohere Rerank model. While this method builds on state-of-the-art developments within the field of information retrieval, there are many other methods and tools we could have chosen, which could impact the context characteristics and model behaviour \citep{wang-etal-2024-rear,katsimpras-paliouras-2024-genra,chen-etal-2024-dense}.
A \textit{comprehensive comparison of different retrieval methods} and their impact on context utilisation would be an interesting direction for future work.

In our creation of \texttt{DRUID}, we ensure to source claims from many different fact-checking sites to increase the representation of our dataset to the entire English-speaking world. However, it is not a uniform distribution, and the amount of context gathered per claim as well as the inter-annotator agreement for the context stances differs across claim sources. This could be due to unintentional cultural biases within our retrieval system or our annotators. Future work could investigate the impact of these \textit{cultural biases in the retrieval process on model output}. \texttt{DRUID}, given its wide distribution of claim and evidence sources, would be an excellent dataset for such an investigation.

\texttt{DRUID} is based on a fact-checking task, meaning that it covers a limited set of domains. This may affect the conclusions based on the dataset, compared to other datasets situated in different domains. Meanwhile, it is worth noting that no dataset is free of this problem -- both CounterFact and ConflictQA have a limited scope by only focusing on WikiData triplets. \texttt{DRUID} is in comparison to these datasets better at covering different domains. Other more realistic datasets would also be limited by the characteristics of the domains they are situated in. For future work it would be interesting to \emph{investigate whether one can control for and disentangle domain effects on context utilisation}, something the \texttt{DRUID} dataset should be useful for.

While we investigate the impact of many characteristics on context utilisation, it is not exhaustive. Future work could look into the impact of \textit{other context characteristics on context utilisation}. For example, our study and dataset omit interesting context characteristics related to propaganda, simplified or manipulated content, anecdotal, mix of languages, multimodality, multi-hop reasoning, preciseness etc. \citep{piskorski-etal-2023-semeval,wan-etal-2021-dqn,jiang-etal-2020-hover,dufour2024ammebalargescalesurveydataset}. These properties would be relevant to study in future work. 

In this work, we study the context utilisation behaviours of Llama 3.1 8B and Pythia 6.9B, two popular LMs used for RAG-evaluation studies.  
With this selection, we represent two families of models and can already reveal great disparities in context utilisation between synthetic and real-world datasets. For future work, it would be interesting to further investigate the context utilisation of \textit{more model families and different model sizes}. All future studies are well-facilitated by the dataset and evaluation framework we introduce in this work.

While we include a comprehensive correlation analysis to identify the dependence between our studied characteristics and context usage, it does not give any information about causality. Future work could include a more \textit{comprehensive causal analysis}. A causal analysis is necessary to fully understand the effects of different context characteristics on context utilisation \citep{feder-etal-2022-causal}. Given that our findings indicate that context utilisation cannot be predicted by one singleton characteristic, there are likely many potential confounders within \texttt{DRUID}, and all real, retrieved text.
Future work on this could take inspiration from the studies by \citet{guicausal}. While our work provides a good starting point for RAG evaluations of context characteristics, our findings show that more work is needed to fully understand the complex behaviours governing context usage.

\section{Ethical Considerations}

Our work concerns the evaluation of RAG-based models on veracity prediction in a real-world setting. In the creation of the dataset, while we tried to maintain representativeness of the real world by including sources of data from different parts of the world, we introduced biases by selecting only English language sources. Consequently, our results only stand for claims and corresponding evidence sentences in English. For the annotation tasks, we do not retain any information about the annotators and pay them a fair wage as determined by the annotation platform. We also informed the annotators about how their data would be used and received their consent. However, for ease of understanding the subject matter and increasing chances of agreement, we screened the annotator pool to only include participants with at least an undergraduate degree, English fluency, no language-related disorders, and UK, US or Irish nationality. While this helped achieve higher-quality annotations, it limits the perspectives embedded in the dataset and may reinforce cultural biases, which we acknowledge as a potential risk. 

Otherwise, we do not foresee any pressing potential risks with this work. We performed foundational research focused on evaluation, which should come with few implications for malicious use, environmental impact, security violations, etc.

\section*{Acknowledgements}
$\begin{array}{l}\includegraphics[width=1cm]{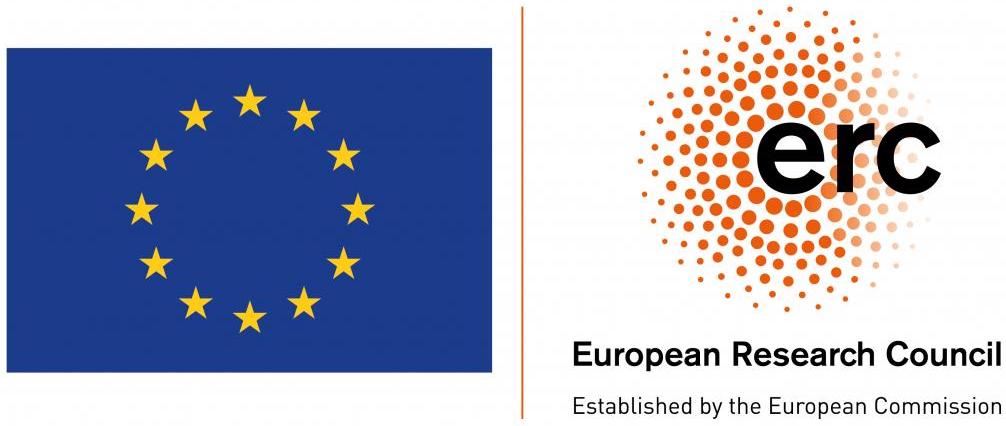} \end{array}$ 
This research was co-funded by the European Union (ERC, ExplainYourself, 101077481), by the Pioneer Centre for AI, DNRF grant number P1, as well as by The Villum Synergy Programme. Views and opinions expressed are however those of the author(s) only and do not necessarily reflect those of the European Union or the European Research Council. Neither the European Union nor the granting authority can be held responsible for them.
The research was also co-funded by the Wallenberg AI, Autonomous Systems and Software Program (WASP) funded by the Knut and Alice Wallenberg Foundation, as well as by WARA Media and Language, also a part of WASP. The computations were enabled by resources provided by the National Academic Infrastructure for Supercomputing in Sweden (NAISS) at Alvis, partially funded by the Swedish Research Council through grant agreement no. 2022-06725. The work was also supported by compute credits from a Cohere For AI Research Grant.
Lastly, we would like to thank the anonymous reviewers for their helpful suggestions.

\bibliography{anthology,custom}

\appendix

\section{Computational resources}

All models are evaluated without fine-tuning on one A40 Nvidia GPU per evaluation. The total computational budget for the evaluation was about 100 GPU hours.

\section{Use of AI assistants}

AI assistants like Copilot and ChatGPT were intermittently used to generate template code and rephrase sentences in the paper, etc. However, no complete paper sections or code scripts have been generated by an AI assistant. All generated content has been inspected and verified by the authors.

\section{DRUID}\label{app:DRUID}

Dataset statistics for \texttt{DRUID+} can be found in \Cref{tab:DRUID+} and statistics for claims with inter-context conflicts can be found in \Cref{tab:DRUID-conflict}. Inter-context conflicts are measured based on the annotated stance of each evidence piece; if we find evidence pieces with different stance (`Refutes' or `Supports') for the same claim, we mark the claim as having conflicting evidence. 

A comparison of \texttt{DRUID} to other fact-checking datasets can be found in \Cref{tab:FCdatasets}. FEVER is marked as \emph{synthetic} and not representative of realistic scenarios for context retrieval as the samples in the dataset have been artificially generated by the following process: 1) take a random sentence from Wikipedia, 2) give this sentence to an annotator and ask them to make up a set of claims based on the sentence, 3) ask annotators to produce additional mutations of the claims and 4) fetch matching contexts mainly from the matching Wikipedia pages. This process is not representative of a realistic use case for context augmentation, as the claims do not match real information needs and the evidence is sourced from nearly perfect Wikipedia page matches. VitaminC and SciFact have been synthesised in a similar fashion, and are therefore also marked as synthetic.

\begin{table}[htbp]
    \centering
    \scriptsize
    \begin{tabular}{lrr}
    \toprule
    Source & \#claims & \#samples \\
    \midrule
    checkyourfact & 300 & 6,653 \\
    climate/health/science.feedback & 293 & 6,983 \\
    factcheckni & 137 & 3,124 \\
    factly & 299 & 6,443 \\
    politifact & 300 & 7,954 \\
    srilanka.factcrescendo & 173 & 4,093 \\
    borderlines & 503 & 3,124 \\
    \midrule
    Total & 2,005 & 48,517 \\
    \bottomrule
    \end{tabular}
    \caption{Statistics for the \texttt{DRUID+} dataset.}
    \label{tab:DRUID+}
\end{table}

\begin{table}[htbp]
    \centering
    \scriptsize
    \begin{tabular}{lrr}
    \toprule
    & & \#confl. \\
    Source & \#claims & claims \\
    \midrule
    checkyourfact & 220 & 112 \\
    climate/health/science.feedback & 220 & 77 \\
    factcheckni & 109 & 25 \\
    factly & 180 & 65 \\
    politifact & 220 & 70 \\
    srilanka.factcrescendo & 156 & 61 \\
    borderlines & 224 & 41 \\
    \midrule
    Total & 1,329 & 451 \\
    \bottomrule
    \end{tabular}
    \caption{Inter-context conflict statistics for the \texttt{DRUID} dataset \citep{xu-etal-2024-knowledge-conflicts}. `\#confl. claims' denotes the number of claims for which we find inter-context conflicts, i.e. conflicting evidence pieces for which at least one evidence piece supports the claim and at least another refutes it.}
    \label{tab:DRUID-conflict}
\end{table}

\begin{table*}[!h]
\centering
\small
\resizebox{\textwidth}{!}{
\small
\begin{tabular}{@{}l|ccccc@{}}
\toprule
\multicolumn{1}{l}{\multirow{2}{*}{\textbf{Dataset}}} & \multicolumn{2}{c}{\textbf{Claim}}                   & \multicolumn{3}{c}{\textbf{Evidence}}                  \\ \cmidrule(l){2-6} 
& \textit{Source} & \multicolumn{1}{c|}{\textit{Type}} & \textit{Sufficient} & \textit{Unleaked} & \textit{Retrieved} \\ \midrule
\multicolumn{1}{l|}{FEVER~\citep{thorne-etal-2018-fever}}          & W               & \multicolumn{1}{c|}{Synthetic}  & \textcolor{blue}{\cmark}   & N/A & \textcolor{blue}{\cmark}  \\
\multicolumn{1}{l|}{VitaminC~\citep{schuster-etal-2021-get}}       & W               & \multicolumn{1}{c|}{Synthetic}  & \textcolor{blue}{\cmark}   & N/A & \textcolor{blue}{\cmark}  \\
\multicolumn{1}{l|}{SciFact~\citep{wadden-etal-2020-fact}}         & S               & \multicolumn{1}{c|}{Synthetic}  & \textcolor{blue}{\cmark}   & N/A & \textcolor{blue}{\cmark}  \\ \midrule
\multicolumn{1}{l|}{Liar-Plus~\citep{alhindi-etal-2018-evidence}}  & FC              & \multicolumn{1}{c|}{Real}      & \textcolor{blue}{\cmark}   & \textcolor{orange}{\xmark} & \textcolor{orange}{\xmark}\\
\multicolumn{1}{l|}{MultiFC~\citep{augenstein-etal-2019-multifc}}  & FC              & \multicolumn{1}{c|}{Real}      & \textcolor{orange}{\xmark} & \textcolor{orange}{\xmark} & \textcolor{blue}{\cmark}  \\
\multicolumn{1}{l|}{WatClaimCheck~\citep{khan-etal-2022-watclaimcheck}}                & FC              & \multicolumn{1}{c|}{Real} & \textcolor{orange}{\xmark} & \textcolor{blue}{\cmark}   & \textcolor{orange}{\xmark}\\
\multicolumn{1}{l|}{ClaimDecomp~\citep{chen-etal-2022-generating}} & FC              & \multicolumn{1}{c|}{Real}      & \textcolor{orange}{\xmark} & \textcolor{blue}{\cmark}   & \textcolor{orange}{\xmark} \\
\multicolumn{1}{l|}{Snopes~\citep{hanselowski-etal-2019-richly}}   & FC              & \multicolumn{1}{c|}{Real}      & \textcolor{orange}{\xmark} & \textcolor{blue}{\cmark}   & \textcolor{orange}{\xmark} \\
\multicolumn{1}{l|}{QABrief~\citep{fan-etal-2020-generating}}      & FC              & \multicolumn{1}{c|}{Real}      & \textcolor{orange}{\xmark} & \textcolor{blue}{\cmark}   & \textcolor{orange}{\xmark} \\
\multicolumn{1}{l|}{CHEF~\citep{hu-etal-2022-chef}}                & FC              & \multicolumn{1}{c|}{Real}      & \textcolor{blue}{\cmark}   & \textcolor{orange}{\xmark} & \textcolor{blue}{\cmark} \\
\multicolumn{1}{l|}{AVeriTeC \citep{averitec}}    & FC              & \multicolumn{1}{c|}{Real}      & \textcolor{blue}{\cmark}   & \textcolor{blue}{\cmark}   & \textcolor{blue}{\cmark}\\ 
\multicolumn{1}{l|}{Factcheck-Bench~\citep{wang-etal-2024-factcheck}}                               & LLM               & \multicolumn{1}{c|}{Real/Synthetic}         & \textcolor{blue}{\cmark}\textcolor{orange}{\xmark}                            & \textcolor{blue}{\cmark}                            & \textcolor{blue}{\cmark}                            \\ \midrule
\multicolumn{1}{l|}{\texttt{DRUID}}                               & W, FC           & \multicolumn{1}{c|}{Real}        & \textcolor{blue}{\cmark}\textcolor{orange}{\xmark} & \textcolor{blue}{\cmark}\textcolor{orange}{\xmark} & \textcolor{blue}{\cmark}  \\ \bottomrule
\end{tabular}
}
\caption{Comparison of related fact-checking datasets. \textit{Source} indicates where the claims are collected from, such as \textbf{W}ikipedia, \textbf{F}act-\textbf{C}hecking articles, \textbf{S}cientific sources or \textbf{LLM} responses. \textit{Type} indicates whether the claims are synthetic or real-world.\textit{Sufficient} indicates whether the evidence can provide sufficient information. \textit{Unleaked} means whether the evidence contains leaks from the future. \textit{Retrieved} denotes whether the dataset involves evidence retrieval instead of relying on pre-retrieved passages e.g.\ the fact-checking article. \textcolor{blue}{\cmark}\textcolor{orange}{\xmark} indicates that both properties can be found and are annotated for.
}
\vspace{-3mm}
\label{tab:FCdatasets}
\end{table*}

\section{Dataset creation}\label{app:dataset}

\subsection{Claim collection}

Different fact-check organisations use different notations for the fact-check verdicts, ranging from `Pants on Fire' to `Inaccurate' \citep{augenstein-etal-2019-multifc}. We only collect claims for which the verdict could be mapped to `True', `False' or `Half-true' (See \Cref{app:dataset}). We collect claims from 7 fact-checking sources, with varying themes and countries of origin:
\begin{itemize}[nosep]
    \item checkyourfact.com
    \item (climatefeedback.org, healthfeedback.org, science.feedback.org)
    \item factcheckni.org
    \item factly.in
    \item politifact.com
    \item srilanka.factcrescendo.com
    \item borderlines from \citet{li-etal-2024-land}
\end{itemize}
The method for sampling the claims is adapted to balance out the precedence of False and US-centric fact-checked claims. To this end, we sample the claims to ensure, to the extent possible, (1) an even distribution across the 7 fact-checking sources, (2) an even distribution across True, False and Half-true claims, and (3) an even distribution of claims posted before and after 2023 (to ensure we also obtain claims and evidence unlikely to be present in the assessed LM's training data). For the sampling, we first prioritise (1), followed by (2) etc. Due to a shortage of some claims, we cannot achieve completely uniform distributions.

We remove all claims that mention a `photo' or a `video' to limit fact-verification to a single media.

\subsection{Claim veracity mappings}\label{app:label-mappings}

We map the claim veracity labels to `True', `False' or `Half-true' as shown in \Cref{tab:label-mappings}.

\begin{table}[!h]
    \centering
    \small
    \begin{tabular}{l p{5cm}}
    \toprule
       Our label  & Incoming label \\
    \midrule
       True  & True \\
       & TRUE \\
       & ACCURATE \\
       & ACCURATE WITH CONSIDERATION \\
       & Correct \\
       & Mostly accurate \\
       & Accurate \\
       Half-true & Half True \\
       & PARTLY TRUE \\
       & Correct But... \\
       & Mostly\_Accurate \\
       & Partially correct \\
       False & False \\
       & FALSE \\
       & MISLEADING \\
       & Misleading \\
       & Inaccurate \\
       & Incorrect, Flawed\_Reasoning \\
       & INACCURATE \\
       & INACCURATE WITH CONSIDERATION \\
    \bottomrule
    \end{tabular}
    \caption{The claim veracity label mappings used for the creation of \texttt{DRUID} and \texttt{DRUID+}. Claims corresponding to verdict labels not listed in the table are dropped.}
    \label{tab:label-mappings}
\end{table}

\subsection{Automated evidence retrieval}

Given a claim, the method is as follows:

\paragraph{Use search engines to search the web for relevant web pages.} Fetch the top 20 search results for the claim from the Google and Bing search engines, respectively.\footnote{We used their respective APIs \url{customsearch.googleapis.com/customsearch/v1} and \url{api.bing.microsoft.com/v7.0/search}. Search results were retrieved in October 2024.} The results are de-duplicated as they may overlap. From this step on, no regard is paid to the search engine ranks, while they are stored for potential future use cases. 

\paragraph{Chunk the content of each web page.} The search results consist of full web pages, for which not all page content can be expected to be immediately useful for the claim verification. Similarly to \citet{diggelmann2020climatefever} we use an extractive approach based on chunking to get concise evidence that fits into the model context window. Each paragraph on the web page forms a chunk if it contains fewer than 200 words. Paragraphs longer than 200 words are split into multiple chunks of up to 200 words. This approach is based on manual tuning and inspection of some retrieved evidence. 

\paragraph{Get reranker scores for the chunks.} The search engines provide a quite coarse filter for relevant information with high recall but low precision. Moreover, the search engines cannot extract the relevant snippets from the search results. To get more precise and accurate retrieved contexts we use a reranker \citep{diggelmann2020climatefever}. Specifically, we use the Cohere Rerank model\footnote{\texttt{rerank-english-v3.0} from \url{https://docs.cohere.com/v2/docs/rerank-2}.} to get reranker scores for each chunk with respect to a claim. 

To avoid claim repeats in the evidence, we also filter out sentences from paragraphs corresponding to $RougeL(sentence,claim)>0.8$ in the chunking step (step 3). Otherwise, the Cohere Rerank model was prone to fetch evidence that more or less only repeated the claim.

\paragraph{Select web pages for evidence retrieval.} For \texttt{DRUID} we have a limited annotation budget and therefore select the four web pages for which we record the maximum reranker chunk scores and collect evidence from each of these. To represent the situation of not having access to fact-checking articles published after the claim was made, we adapt this selection to collect at minimum two pieces of evidence posted before the publication of the claim. This way, we ensure that at least half of the dataset contains unleaked information. For \texttt{DRUID+} we select all webpage search results for the evidence collection.

\paragraph{Collect evidence from the selected web pages.} Collect an evidence piece from each of the web pages selected in the previous step. This is done by aggregating the three top-ranked chunks from the web page via simple concatenation. If necessary, the number of chunks is decreased to ensure that no evidence piece is longer than 300 words. As a result, we have several pieces of evidence per claim, each representative of one web page.

\section{Additional dataset details}
\label{app:other_datasets}

\subsection{CounterFact}

\begin{table}[htbp]
    \centering
    \small
    \begin{tabular}{lp{4cm}}
    \toprule
    Column & Value \\
    \midrule
    Claim & Geoffrey Hinton is employed by BBC. \\
    Verdict & False \\
    \midrule
    Evidence \#1 & Geoffrey Hinton is employed by BBC. \\
    Relevant & True \\
    Evidence stance & Supports \\
    \midrule
    Evidence \#2 & Geoffrey Hinton is employed by Google. \\
    Relevant & True \\
    Evidence stance & Refutes \\
    \bottomrule
    \end{tabular}
    \caption{A sample from CounterFact that has been recast to match the format of \texttt{DRUID}.}
    \label{tab:counterfact-recast-example}
\end{table}

The CounterFact dataset referred to in this paper has been developed by \citet{ortu-etal-2024-competition} to study context usage under knowledge conflicts. It contains 10,000 samples based on fact triplets from WikiData. An example of a sample from the CounterFact dataset is ``Redefine: Geoffrey Hinton is employed by BBC. Geoffrey Hinton is employed by''. A knowledge conflict is induced by the replacement of the correct answer (\emph{Google}) with \emph{BBC} in the context. We use the CounterFact split based on Pythia 6.9B.

To ensure alignment between the investigations for CounterFact and \texttt{DRUID} we first recast the CounterFact samples to a format that aligns with the \texttt{DRUID} dataset. This is exemplified in \Cref{tab:counterfact-recast-example}. The queries are recast to claims and we retain both the new knowledge conflicting context as well as the original correct context as evidence. By virtue of the synthetic nature of the dataset, we know beforehand that all claims are incorrect and that the new contexts support the claims. The opposite holds for the original, correct, contexts. We also know that all contexts are relevant to the claims.

\subsection{ConflictQA}

\begin{table*}[htbp]
    \centering
    \small
    \begin{tabular}{lp{10cm}}
    \toprule
    Column & Value \\
    \midrule
    Memory answer & George Rankin is a lawyer. \\
    Memory aligned evidence & George Rankin graduated from Harvard Law School in 2005 and has been practicing law for the past 15 years. He is a member of the American Bar Association and has been recognized as a leading lawyer in the field of intellectual property law by several prestigious legal publications. In addition, he currently serves as a partner at one of the top law firms in the country. \\
    Counter memory aligned evidence & George Rankin Major General George James Rankin, (1 May 1887 - 28 December 1957) was an Australian soldier and politician. He served in both the House of Representatives and the Senate, representing the Country Party of Australia. Rankin was born at Bamawm, Victoria, the tenth child of Irish farmer James Rankin and Sarah, né Gallagher. He attended the local state school and became a farmer. In 1907, he joined the Militia, and was commissioned in the 9th Light Horse Regiment in 1909. He married Annie Isabella Oliver at Rochester, Victoria on 7 July 1912. In 1914, he was appointed a \\
    \bottomrule
    \end{tabular}
    \caption{A sample from the ConflictQA dataset.}
    \label{tab:conflictQA-example}
\end{table*}

We also inspect the ConflictQA dataset developed by \citet{xie2024knowledgeconflict}. The dataset contains `memory answers' from an LM (based on its parametric memory) to prompts from PopQA (also based on WikiData fact triplets) together with `counter answers' generated by an LLM instructed to produce an answer that conflicts with the model answer. Each entry also contains corresponding evidence, one that is `parametric memory aligned' and another that is `counter memory aligned'. These evidence pieces have been generated or sourced from Wikipedia/human annotation. We use the ConflictQA split based on Llama 2 7B. An example from the ConflictQA dataset can be found in \Cref{tab:conflictQA-example}.

We use the `memory answer' (generated by Llama 2 7B) as the claim and the `parametric memory aligned evidence' and `counter memory aligned evidence' as supporting and refuting evidence corresponding to the claim.

The generated origin of the evidence is revealed at multiple instances (see below). Moreover, the generated evidence is many times directly on point which cannot expected to be found in real-world scenarios.

\section{Attention manipulation results on CounterFact after Reformatting}\label{app:manipulation-results-after-reformat}

\begin{figure}[!h]
    \centering
    \includegraphics[width=\linewidth,trim={0 0 0 1.2cm},clip]{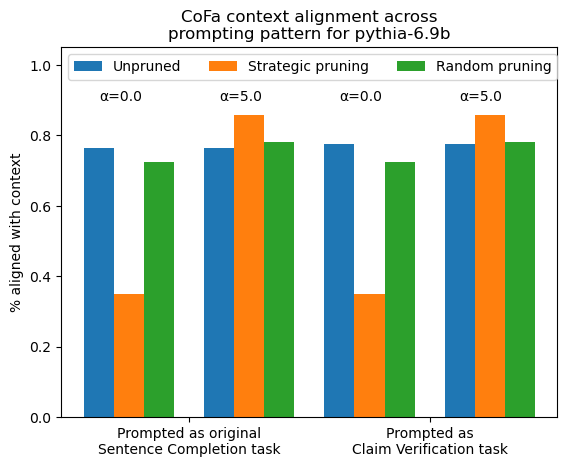}
    \caption{The results of pruning attention heads in Pythia for the original sentence completion task and for when the task has been recast to a claim verification task.}
    \label{fig:attention-manipulation-results}
\end{figure}

\Cref{fig:attention-manipulation-results} shows the results of pruning attention heads in Pythia for the original sentence completion task as studied by \citet{ortu-etal-2024-competition} compared to the same approach but recast to a claim verification task. The effects of attention head pruning are largely unaffected by the reformatting to a claim verification task, showing that LMs can be interpreted and manipulated for the claim verification task just as well as for the sentence completion task.

\section{Jaccard similarity to proxy claim-evidence similarity}\label{app:jaccard}

We use Jaccard similarity to proxy claim-evidence similarity. This is measured as follows.
\begin{equation}
    J(C, E) = \frac{|W(C) \cap W(E)|}{|W(C) \cup W(E)|} 
\end{equation}
$W$ denotes the set of unique words, lowercased and ignoring punctuation or special characters like `-', found in a claim $C$ or evidence $E$. 

\section{Cohere: Refers to external source} \label{app:cohere-detect-references}

The prompt used for the detection of references to external sources with Cohere Command R+ is as follows: ``Does the following text refer to an external source or not? Admissible external sources are for example `a study', `[1]', `the BBC', a news channel etc. Answer with a `Yes' or `No'.\textbackslash n\textbackslash nText: <text>''.

\section{Context characteristics}\label{app:characteristics}
The full statistics on the context characteristics for all datasets considered can be found in \Cref{tab:property-statistics-relevant,tab:property-statistics-stance,tab:property-statistics-general}.

\begin{table}[!t]
    \centering
    \small
    \begin{tabular}{llll}
    \toprule
    Relevant & CounterFact & ConflictQA & \texttt{DRUID} \\
    \midrule
    True & 20,000 & 16,046 & 5,399 \\
    False & 0 & 0 & 91 \\
    \bottomrule
    \end{tabular}
    \caption{Evidence relevance for each of the investigated datasets.}
    \label{tab:property-statistics-relevant}
\end{table}

\begin{table}[!t]
    \centering
    \small
    \begin{tabular}{lrrr}
    \toprule
    Evidence stance & CounterFact & ConflictQA & \texttt{DRUID} \\
    \midrule
    refutes & 10,000 & 8,023 & 1,760 \\
    insufficient & 0 & 0 & 2,730 \\
    \: -refutes & 0 & 0 & 557 \\
    \: -contradictory & 0 & 0 & 410 \\
    \: -neutral & 0 & 0 & 1,078 \\
    \: -supports & 0 & 0 & 685 \\
    supports & 10,000 & 8,023 & 909 \\
    \bottomrule
    \end{tabular}
    \caption{Evidence stance for each of the investigated datasets.}
    \label{tab:property-statistics-stance}
\end{table}

\begin{table*}[!h]
    \centering
    \scriptsize
    \begin{tabular}{lrrrr}
    \toprule
    \textbf{Property} & \textbf{CounterFact} & \textbf{ConflictQA} & \textbf{\texttt{DRUID+}} & \textbf{\texttt{DRUID}} \\
    \midrule
    \textbf{Claim-evidence similarity} \\
    Jaccard similarity & $\mathbf{0.89 \pm 0.12}$ & $0.09 \pm 0.04$ & $0.09 \pm 0.06$ & $0.12 \pm 0.08$ \\
    Claim-evidence overlap & $\mathbf{0.93 \pm 0.08}$ & $0.76 \pm 0.23$ & $0.58 \pm 0.25$ & $0.66 \pm 0.26$ \\
    Repeats claim (\%) & (50.00) & 5.55 & 1.25 & 4.57 \\ 
    \textbf{Difficult to understand} \\
    Flesch reading ease score & $61.65 \pm 22.50$ & $56.25 \pm 12.00$ & $53.17 \pm 24.74$ & $53.54 \pm 16.60$ \\
    Claim length & $44.70 \pm 11.90$ & $43.64 \pm 15.69$ & $\mathbf{84.08 \pm 46.37}$ & $\mathbf{89.25 \pm 46.15}$ \\
    Evidence length & $\mathbf{44.63 \pm 11.90}$ & $570.46 \pm 158.61$ & $775.64 \pm 407.40$ & $745.39 \pm 406.33$ \\
    Llama: Perplexity & $\mathbf{172.94 \pm 537.58}$ & $7.55\pm5.00$ & $17.22\pm124.59$ & $16.08 \pm 43.82$ \\
    Pythia: Perplexity & $\mathbf{113.43 \pm 1030.10}$ & $9.29\pm4.51$ & $19.35\pm122.17$ & $18.13 \pm 32.55$ \\
    \textbf{Implicit} \\
    Claim entity overlap & $0.75 \pm 0.27$ & $0.69 \pm 0.36$ & $0.46 \pm 0.39$ & $0.56 \pm 0.40$ \\
    \textbf{Refers external source} \\
    Detection by LLM (\%) & (0.00) & 27.37 & - & $\mathbf{40.55}$ \\
    \textbf{Unreliable} \\
    Unreliable source (\%) & - & - & 5.00 & 3.50 \\
    \textbf{Uncertain} \\
    Contains hedging (\%) & $\mathbf{0.06}$ & 15.34 & 36.61 & 36.54 \\
    Contains hedging discourse (\%) & $\mathbf{0.03}$ & 40.29 & 48.00 & 52.33 \\
    \textbf{Additional properties} \\
    Contains `True' & 0.00 & 1.99 & 2.57 & 4.06 \\
    Contains `False' & 0.00 & 0.10 & 4.27 & 9.02 \\
    Fact-check source (\%) & - & - & 14.41 & $\mathbf{41.44}$ \\
    Gold source (\%) & - & - &  4.13 & 17.21 \\
    Pub. after claim (\%) & - & - & 53.37 & 50.26 \\
    \midrule
    \textbf{Total instances} & 20,000 & 16,046 & 48,517 & 5,490 \\
    \bottomrule
    \end{tabular}
    \caption{Statistics for the context characteristics in CounterFact \citep{yu-etal-2023-characterizing}, ConflictQA \citep{xie2024knowledgeconflict} and \texttt{DRUID} datasets. The characteristics and their detection are described in \Cref{sec:context-characteristics,sec:characteristics-detection}, respectively. The values indicate the mean $\pm$ the standard deviation or the percentage of claim-evidence samples affected, denoted with (\%). The LLM-detected properties for CounterFact indicated with a (parenthesis) were not LLM detected but automatically detected for each sample, leveraging its synthetic nature. Outliers are marked in \textbf{bold}.}
    \label{tab:property-statistics-general}
\end{table*}

\begin{table}[!h]
    \centering
    \scriptsize
    \begin{tabular}{lllr}
    \toprule
        Dataset & Model & Prompt & ACU \\
    \midrule
        CounterFact & Llama & \textbf{3-shot} & \textbf{0.74} \\
        & & 0-shot & 0.57 \\
        & Pythia & 3-shot & -0.05 \\
        & & \textbf{0-shot} & \textbf{0.32} \\
        ConflictQA & Llama & \textbf{3-shot} & \textbf{0.71} \\
        & & 0-shot & 0.56 \\
        & Pythia & \textbf{3-shot} & \textbf{0.55} \\
        & & 0-shot & 0.50 \\
        \texttt{DRUID} & Llama & \textbf{3-shot} & \textbf{0.84} \\
        & & 0-shot & 0.34 \\
        & Pythia & \textbf{3-shot} & \textbf{0.25} \\
        & & 0-shot & 0.15 \\
    \bottomrule
    \end{tabular}
    \caption{The averaged ACU scores across all evidence stances for each dataset, model and prompt. The better performing prompt and corresponding ACU value is marked in \textbf{bold}.}
    \label{tab:cu-mean-prompts}
\end{table}

\section{Prompts}\label{app:prompts}

For each mode and model, we manually tune a prompt on 390 samples from \texttt{DRUID} to maximise context usage, using the balanced mean absolute error\footnote{We used sci-kit learn's \href{https://scikit-learn.org/1.5/modules/generated/sklearn.metrics.mean_absolute_error.html}{\texttt{mean\_absolute\_error}} with sample weights given by \href{https://scikit-learn.org/1.5/modules/generated/sklearn.utils.class_weight.compute_sample_weight.html}{\texttt{compute\_sample\_weight}} for the `balanced' setting.} as the objective function to be minimised. For mode (1) the gold labels are given by the claim veracity and for (2) the annotated evidence stances. We experimented with around 21 different prompts (0-shot, 2-shot and 3-shot) in total. The best-performing prompts were found to be 3-shot. All prompts request the model to say whether the claim, with or without evidence, is `True' or `False'. Moreover, the model is instructed to respond `None' if it is uncertain or cannot answer.

The tuned prompts used to evaluate Llama and Pythia can be found in \Cref{tab:prompts-claim,tab:prompts-claim-evidence,tab:prompts-claim-evidence-part-2}. The tables also list the prompts used for the 0-shot experiments described in the appendix.

\begin{table*}[htbp]
    \centering
    \scriptsize
    \begin{tabular}{p{1cm}lp{12cm}}
    \toprule
    Model & \#shots & Prompt \\
    \midrule
    Llama & 3 & Are the following claims True or False? Answer None if you are not sure or cannot answer.
    \bigbreak
    Claimant: Joe Biden
    
    Claim: "“One quarter” of today’s \$31.4 trillion federal debt “was accumulated in the four years of my predecessor,” Donald Trump."
    
    Answer: True
    \bigbreak
    Claimant: Viral post
    
    Claim: "the new coronavirus has HIV proteins that indicate it was genetically modified in a laboratory."
    
    Answer: False
    \bigbreak
    Claimant: Sara Daniels
    
    Claim: "Blackpink released the single 'You me too' in 2026."
    
    Answer: None
    \bigbreak
    Claimant: <claimant>
    
    Claim: "<claim>"
    
    Answer: \\
    \midrule
    Pythia & 3 & Are the following claims True or False? Answer None if you are not sure or cannot answer.
    \bigbreak
    Claimant: Joe Biden
    
    Claim: "“One quarter” of today’s \$31.4 trillion federal debt “was accumulated in the four years of my predecessor,” Donald Trump."
    
    Answer: True
    \bigbreak
    Claimant: Viral post
    
    Claim: "5G causes cancer."
    
    Answer: False
    \bigbreak
    Claimant: Sara Daniels
    
    Claim: "Blackpink released the single 'You me too' in 2026."
    
    Answer: None
    \bigbreak
    Claimant: <claimant>
    
    Claim: "<claim>"
    
    Answer:\\
    \midrule
    Llama \& Pythia & 0 & Is the following claim True or False? Answer None if you are not sure or cannot answer.
     \bigbreak
     Claimant: <claimant>
     
     Claim: "<claim>"
     
     Answer: \\
    \bottomrule
    \end{tabular}
    \caption{Prompts used to evaluate Pythia and Llama in a claim-only setting. Values in <brackets> are replaced by the actual entry for the evaluated sample. For CounterFact and ConflictQA we omit the `Claimant' lines as there are no claimant entries for these datasets.}
    \label{tab:prompts-claim}
\end{table*}

\begin{table*}[htbp]
    \centering
    \scriptsize
    \begin{tabular}{llp{12cm}}
    \toprule
    Model & \#shots & Prompt \\
    \midrule
    Llama & 3 & Here are some claims and corresponding evidence. Does the evidence Support or Refute the claim? Answer None if there is not enough information in the evidence to decide.
    \bigbreak
    Claimant: Joe Biden
    
    Claim: "“One quarter” of today’s \$31.4 trillion federal debt “was accumulated in the four years of my predecessor,” Donald Trump."
    
    Evidence: "Biden’s number is accurate; about one-fourth of the total debt incurred to date came on Trump’s watch. However, assigning debt to a particular president is tricky, because so much of the spending was approved by decades-old, bipartisan legislation that set the parameters for Social Security and Medicare. A different calculation shows more debt stemming from former President Barack Obama, with whom Biden served as vice president."
    
    Answer: Support
    \bigbreak
    Claimant: Viral post
    
    Claim: "the new coronavirus has HIV proteins that indicate it was genetically modified in a laboratory."
    Evidence: "Microbiologists say the spike proteins found in the new coronavirus are different from the ones found in HIV. [...] There is no evidence to suggest the coronavirus was genetically modified."
    
    Answer: Refute
    \bigbreak
    Claimant: Sara Daniels
    
    Claim: "Blackpink released the single 'You me too' in 2026."
    
    Evidence: "Blackpink released their album 'Born Pink' in 2022."
    
    Answer: None
    \bigbreak
    Claimant: <claimant>
    
    Claim: "<claim>"
    
    Evidence: "<evidence>"
    
    Answer: \\
    \bottomrule
    \end{tabular}
    \caption{Prompts used to evaluate Pythia and Llama in a setting with provided claim and evidence. Values in <brackets> are replaced by the actual entry for the evaluated sample. For CounterFact and ConflictQA we omit the `Claimant' lines as there are no claimant entries for these datasets.}
    \label{tab:prompts-claim-evidence}
\end{table*}

\begin{table*}[htbp]
    \centering
    \scriptsize
    \begin{tabular}{llp{12cm}}
    \toprule
    Model & \#shots & Prompt \\
    \midrule
    Pythia & 3 & Are the claims True or False based on the accompanying evidence? If you are not sure or cannot answer, say None.
    \bigbreak
    Claimant: Joe Biden
    
    Claim: "“One quarter” of today’s \$31.4 trillion federal debt “was accumulated in the four years of my predecessor,” Donald Trump."
    
    Evidence: "Biden’s number is accurate; about one-fourth of the total debt incurred to date came on Trump’s watch. However, assigning debt to a particular president is tricky, because so much of the spending was approved by decades-old, bipartisan legislation that set the parameters for Social Security and Medicare. A different calculation shows more debt stemming from former President Barack Obama, with whom Biden served as vice president."
    
    Answer: True
    \bigbreak
    Claimant: Viral post
    
    Claim: "the new coronavirus has HIV proteins that indicate it was genetically modified in a laboratory."
    Evidence: "Microbiologists say the spike proteins found in the new coronavirus are different from the ones found in HIV. [...] There is no evidence to suggest the coronavirus was genetically modified."
    
    Answer: False
    \bigbreak
    Claimant: Sara Daniels
    
    Claim: "Blackpink released the single 'You me too' in 2026."
    
    Evidence: "Blackpink released their album 'Born Pink' in 2022."
    
    Answer: None
    \bigbreak
    Claimant: <claimant>
    
    Claim: "<claim>"
    
    Evidence: "<evidence>"
    
    Answer:\\
    \midrule
    Llama, Pythia & 0 & Based on the provided evidence, is the claim True or False? If you are not sure or cannot answer, say None.
     \bigbreak
     Claimant: <claimant>
     
     Claim: "<claim>"
     
     Evidence: "<evidence>"
     
     Answer:\\
    \bottomrule
    \end{tabular}
    \caption{Prompts used to evaluate Pythia and Llama in a setting with provided claim and evidence. Values in <brackets> are replaced by the actual entry for the evaluated sample. For CounterFact and ConflictQA we omit the `Claimant' lines as there are no claimant entries for these datasets.}
    \label{tab:prompts-claim-evidence-part-2}
\end{table*}

\section{Additional context usage results} \label{app:additional-cu-results}

\begin{figure*}[!h]
    \centering
    \begin{subfigure}[t]{0.5\textwidth}
        \includegraphics[width=\linewidth]{figures/context_usage/aggregated.pdf}
        \caption{3-shot}
        \label{fig:cu-agg-again}
    \end{subfigure}%
    \begin{subfigure}[t]{0.5\textwidth}
        \includegraphics[width=1\linewidth]{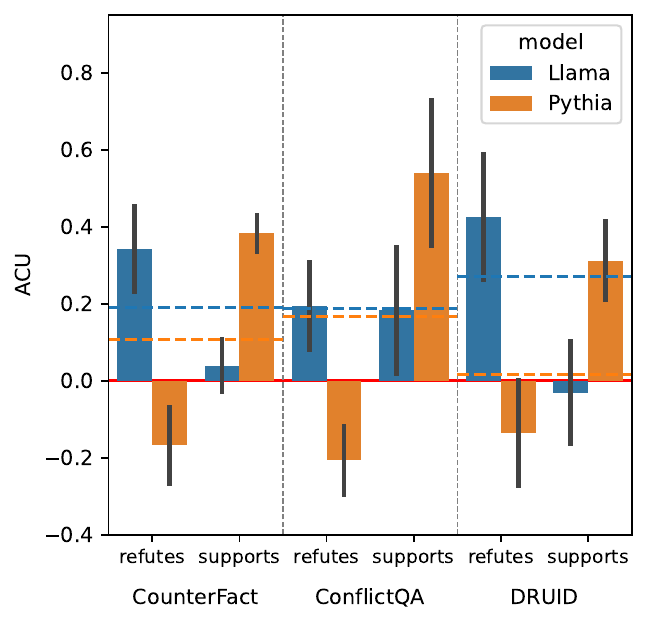}
        \caption{0-shot}
    \label{fig:cu-agg-zero}
    \end{subfigure}
    \begin{subfigure}[t]{0.5\textwidth}
        \includegraphics[width=\linewidth]{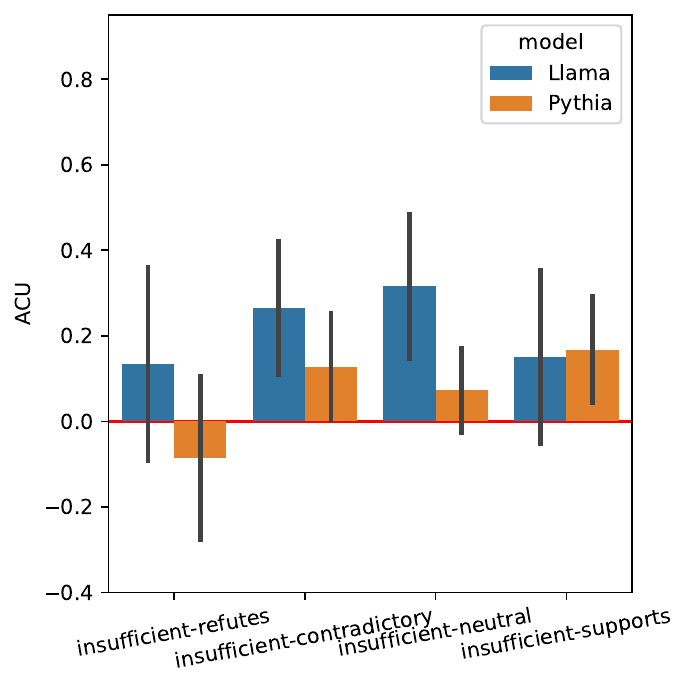}
        \caption{3-shot}
        \label{fig:cu-agg-insuff}
    \end{subfigure}%
    \begin{subfigure}[t]{0.5\textwidth}
        \includegraphics[width=\linewidth]{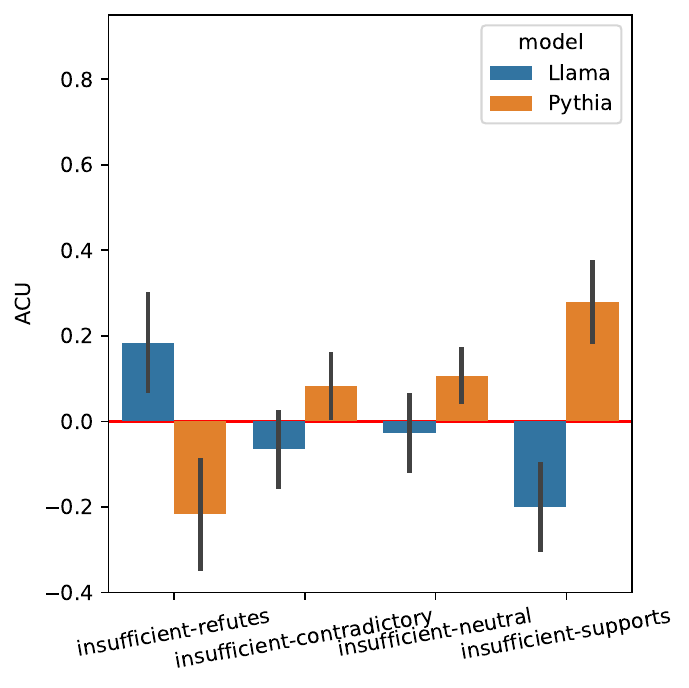}
        \caption{0-shot}
    \label{fig:cu-agg-insuff-zero}
    \end{subfigure}
    \caption{Accumulated context usage (\Cref{eq:acu}) each model on sufficient evidence from all datasets (\Cref{fig:cu-agg-again,fig:cu-agg-zero}) and insufficient evidence from \texttt{DRUID} (\Cref{fig:cu-agg-insuff,fig:cu-agg-insuff-zero}) under different prompts. \Cref{fig:cu-agg} is included again for comparison (\Cref{fig:cu-agg-again}). The error bars indicate the standard deviation. The maximum and minimum context usage value possible is 3 and -3, respectively. Values under the red line indicate `context-repulsion'.}
    \label{fig:cu-agg-all}
\end{figure*}

Some cherry- and lemon-picked samples from the investigated datasets and corresponding model predictions can be found in \Cref{tab:cherry-samples-counterfact,tab:lemon-samples-counterfact,tab:cherry-samples-conflictqa,tab:lemon-samples-conflictqa,tab:cherry-samples-druid,tab:lemon-samples-druid}.
\begin{table*}[h]
    \centering
    \scriptsize
    \begin{tabular}{llll}
    \toprule
       \textbf{Dataset} & \multicolumn{3}{l}{CounterFact} \\
    \textbf{Claim} & \multicolumn{3}{l}{Danish Outdoor Council is headquartered in Perth.} \\
    \textbf{Evidence} & \multicolumn{3}{l}{Danish Outdoor Council is headquartered in Copenhagen.} \\
    \textbf{Evidence stance} & \multicolumn{3}{l}{refutes} \\
    $\bm{\mathrm{ACU}_{\mathrm{Llama}}}$ & \multicolumn{3}{l}{1.51} \\
    $P_{\mathrm{Llama}}(\mathrm{False}|C)$ & 0.69 &
    $P_{\mathrm{Llama}}(\mathrm{False}|C, E)$ & 0.84 \\
    $P_{\mathrm{Llama}}(\mathrm{None}|C)$ & 0.17 &
    $P_{\mathrm{Llama}}(\mathrm{None}|C, E)$ & 0.15 \\
    $P_{\mathrm{Llama}}(\mathrm{True}|C)$ & 0.14 &
    $P_{\mathrm{Llama}}(\mathrm{True}|C, E)$ & 0.01 \\
    $\bm{\mathrm{ACU}_{\mathrm{Pythia}}}$ & \multicolumn{3}{l}{0.13} \\
    $P_{\mathrm{Pythia}}(\mathrm{False}|C)$ & 0.36 &
    $P_{\mathrm{Pythia}}(\mathrm{False}|C, E)$ & 0.34 \\
    $P_{\mathrm{Pythia}}(\mathrm{None}|C)$ & 0.03 &
    $P_{\mathrm{Pythia}}(\mathrm{None}|C, E)$ & 0.26 \\
    $P_{\mathrm{Pythia}}(\mathrm{True}|C)$ & 0.58 &
    $P_{\mathrm{Pythia}}(\mathrm{True}|C, E)$ & 0.34 \\
    \midrule
    \textbf{Dataset} & \multicolumn{3}{l}{CounterFact} \\
    \textbf{Claim} & \multicolumn{3}{l}{Yahoo! Screen is owned by Sony.} \\
    \textbf{Evidence} & \multicolumn{3}{l}{Yahoo! Screen is owned by Sony.} \\
    \textbf{Evidence stance} & \multicolumn{3}{l}{supports} \\
    $\bm{\mathrm{ACU}_{\mathrm{Llama}}}$ & \multicolumn{3}{l}{1.77} \\
    $P_{\mathrm{Llama}}(\mathrm{False}|C)$ & 0.65 &
    $P_{\mathrm{Llama}}(\mathrm{False}|C, E)$ & 0.05 \\
    $P_{\mathrm{Llama}}(\mathrm{None}|C)$ & 0.10 &
    $P_{\mathrm{Llama}}(\mathrm{None}|C, E)$ & 0.09 \\
    $P_{\mathrm{Llama}}(\mathrm{True}|C)$ & 0.25 &
    $P_{\mathrm{Llama}}(\mathrm{True}|C, E)$ & 0.84 \\
    $\bm{\mathrm{ACU}_{\mathrm{Pythia}}}$ & \multicolumn{3}{l}{0.16} \\
    $P_{\mathrm{Pythia}}(\mathrm{False}|C)$ & 0.51 &
    $P_{\mathrm{Pythia}}(\mathrm{False}|C, E)$ & 0.12 \\
    $P_{\mathrm{Pythia}}(\mathrm{None}|C)$ & 0.04 &
    $P_{\mathrm{Pythia}}(\mathrm{None}|C, E)$ & 0.47 \\
    $P_{\mathrm{Pythia}}(\mathrm{True}|C)$ & 0.42 &
    $P_{\mathrm{Pythia}}(\mathrm{True}|C, E)$ & 0.36 \\
    \midrule
    \textbf{Dataset} & \multicolumn{3}{l}{CounterFact} \\
    \textbf{Claim} & \multicolumn{3}{l}{The Voice debuted on CNN.} \\
    \textbf{Evidence} & \multicolumn{3}{l}{The Voice debuted on CNN.} \\
    \textbf{Evidence stance} & \multicolumn{3}{l}{supports} \\
    $\bm{\mathrm{ACU}_{\mathrm{Llama}}}$ & \multicolumn{3}{l}{1.44} \\
    $P_{\mathrm{Llama}}(\mathrm{False}|C)$ & 0.58 &
    $P_{\mathrm{Llama}}(\mathrm{False}|C, E)$ & 0.09 \\
    $P_{\mathrm{Llama}}(\mathrm{None}|C)$ & 0.13 &
    $P_{\mathrm{Llama}}(\mathrm{None}|C, E)$ & 0.17 \\
    $P_{\mathrm{Llama}}(\mathrm{True}|C)$ & 0.28 &
    $P_{\mathrm{Llama}}(\mathrm{True}|C, E)$ & 0.73 \\
    $\bm{\mathrm{ACU}_{\mathrm{Pythia}}}$ & \multicolumn{3}{l}{1.01} \\
    $P_{\mathrm{Pythia}}(\mathrm{False}|C)$ & 0.43 &
    $P_{\mathrm{Pythia}}(\mathrm{False}|C, E)$ & 0.12 \\
    $P_{\mathrm{Pythia}}(\mathrm{None}|C)$ & 0.09 &
    $P_{\mathrm{Pythia}}(\mathrm{None}|C, E)$ & 0.19 \\
    $P_{\mathrm{Pythia}}(\mathrm{True}|C)$ & 0.45 &
    $P_{\mathrm{Pythia}}(\mathrm{True}|C, E)$ & 0.66 \\
   \bottomrule
    \end{tabular}
    \caption{Cherry-picked ACU samples for Llama and/or Pythia on CounterFact.}
    \label{tab:cherry-samples-counterfact}
\end{table*}
\begin{table*}[!h]
    \centering
    \scriptsize
    \begin{tabular}{llll}
    \toprule
    \textbf{Dataset} & \multicolumn{3}{l}{CounterFact} \\
    \textbf{Claim} & \multicolumn{3}{l}{Satchel Paige is a professional basketball.} \\
    \textbf{Evidence} & \multicolumn{3}{l}{Satchel Paige is a professional baseball.} \\
    \textbf{Evidence stance} & \multicolumn{3}{l}{refutes} \\
    $\bm{\mathrm{ACU}_{\mathrm{Llama}}}$ & \multicolumn{3}{l}{0.04} \\
    $P_{\mathrm{Llama}}(\mathrm{False}|C)$ & 0.75 &
    $P_{\mathrm{Llama}}(\mathrm{False}|C, E)$ & 0.55 \\
    $P_{\mathrm{Llama}}(\mathrm{None}|C)$ & 0.10 &
    $P_{\mathrm{Llama}}(\mathrm{None}|C, E)$ & 0.39 \\
    $P_{\mathrm{Llama}}(\mathrm{True}|C)$ & 0.14 &
    $P_{\mathrm{Llama}}(\mathrm{True}|C, E)$ & 0.05 \\
    $\bm{\mathrm{ACU}_{\mathrm{Pythia}}}$ & \multicolumn{3}{l}{-0.14} \\
    $P_{\mathrm{Pythia}}(\mathrm{False}|C)$ & 0.46 &
    $P_{\mathrm{Pythia}}(\mathrm{False}|C, E)$ & 0.35 \\
    $P_{\mathrm{Pythia}}(\mathrm{None}|C)$ & 0.04 &
    $P_{\mathrm{Pythia}}(\mathrm{None}|C, E)$ & 0.27 \\
    $P_{\mathrm{Pythia}}(\mathrm{True}|C)$ & 0.47 &
    $P_{\mathrm{Pythia}}(\mathrm{True}|C, E)$ & 0.31 \\
    \midrule
    \textbf{Dataset} & \multicolumn{3}{l}{CounterFact} \\
    \textbf{Claim} & \multicolumn{3}{l}{Honda SFX, produced by Airbus.} \\
    \textbf{Evidence} & \multicolumn{3}{l}{Honda SFX, produced by Honda.} \\
    \textbf{Evidence stance} & \multicolumn{3}{l}{refutes} \\
    $\bm{\mathrm{ACU}_{\mathrm{Llama}}}$ & \multicolumn{3}{l}{-0.73} \\
    $P_{\mathrm{Llama}}(\mathrm{False}|C)$ & 0.64 &
    $P_{\mathrm{Llama}}(\mathrm{False}|C, E)$ & 0.37 \\
    $P_{\mathrm{Llama}}(\mathrm{None}|C)$ & 0.16 &
    $P_{\mathrm{Llama}}(\mathrm{None}|C, E)$ & 0.42 \\
    $P_{\mathrm{Llama}}(\mathrm{True}|C)$ & 0.20 &
    $P_{\mathrm{Llama}}(\mathrm{True}|C, E)$ & 0.20 \\
    $\bm{\mathrm{ACU}_{\mathrm{Pythia}}}$ & \multicolumn{3}{l}{-0.73} \\
    $P_{\mathrm{Pythia}}(\mathrm{False}|C)$ & 0.44 &
    $P_{\mathrm{Pythia}}(\mathrm{False}|C, E)$ & 0.18 \\
    $P_{\mathrm{Pythia}}(\mathrm{None}|C)$ & 0.14 &
    $P_{\mathrm{Pythia}}(\mathrm{None}|C, E)$ & 0.52 \\
    $P_{\mathrm{Pythia}}(\mathrm{True}|C)$ & 0.36 &
    $P_{\mathrm{Pythia}}(\mathrm{True}|C, E)$ & 0.25 \\
    \midrule
    \textbf{Dataset} & \multicolumn{3}{l}{CounterFact} \\
    \textbf{Claim} & \multicolumn{3}{l}{iPad, developed by Douglas.} \\
    \textbf{Evidence} & \multicolumn{3}{l}{iPad, developed by Douglas.} \\
    \textbf{Evidence stance} & supports \\
    $\bm{\mathrm{ACU}_{\mathrm{Llama}}}$ & \multicolumn{3}{l}{0.44} \\
    $P_{\mathrm{Llama}}(\mathrm{False}|C)$ & 0.49 &
    $P_{\mathrm{Llama}}(\mathrm{False}|C, E)$ & 0.14 \\
    $P_{\mathrm{Llama}}(\mathrm{None}|C)$ & 0.17 &
    $P_{\mathrm{Llama}}(\mathrm{None}|C, E)$ & 0.46 \\
    $P_{\mathrm{Llama}}(\mathrm{True}|C)$ & 0.33 &
    $P_{\mathrm{Llama}}(\mathrm{True}|C, E)$ & 0.38 \\
    $\bm{\mathrm{ACU}_{\mathrm{Pythia}}}$ & \multicolumn{3}{l}{-0.10} \\
    $P_{\mathrm{Pythia}}(\mathrm{False}|C)$ & 0.40 &
    $P_{\mathrm{Pythia}}(\mathrm{False}|C, E)$ & 0.10 \\
    $P_{\mathrm{Pythia}}(\mathrm{None}|C)$ & 0.15 &
    $P_{\mathrm{Pythia}}(\mathrm{None}|C, E)$ & 0.60 \\
    $P_{\mathrm{Pythia}}(\mathrm{True}|C)$ & 0.36 &
    $P_{\mathrm{Pythia}}(\mathrm{True}|C, E)$ & 0.24 \\
   \bottomrule
    \end{tabular}
    \caption{Lemon-picked ACU samples for Llama and/or Pythia on CounterFact.}
    \label{tab:lemon-samples-counterfact}
\end{table*}
\begin{table*}[!h]
    \centering
    \scriptsize
    \begin{tabular}{llll}
    \toprule
     \textbf{Dataset} & \multicolumn{3}{l}{ConflictQA} \\
    \textbf{Claim} & \multicolumn{3}{l}{The screenwriter for Highway was Imtiaz Ali.} \\
    \textbf{Evidence} & \multicolumn{3}{p{12cm}}{Highway is a 2014 Indian Hindi-language road drama film written and directed by Imtiaz Ali and produced by Sajid Nadiadwala. The film stars Alia Bhatt and Randeep Hooda. Screened in the Panorama section of the 2014 Berlin International Film Festival, the film released worldwide on 21 February 2014. The film is based on the episode of the same name from the Zee TV anthology series Rishtey, starring Aditya Srivastava and Kartika Rane, which was also written and directed by Imtiaz Ali. It tells the story of a girl (Alia Bhatt) who, for reasons later revealed, discovers freedom after being kidnapped.} \\
    \textbf{Evidence stance} & \multicolumn{3}{l}{supports} \\
    $\bm{\mathrm{ACU}_{\mathrm{Llama}}}$ & \multicolumn{3}{l}{1.94} \\
    $P_{\mathrm{Llama}}(\mathrm{False}|C)$ & 0.37 &
    $P_{\mathrm{Llama}}(\mathrm{False}|C, E)$ & 0.03 \\
    $P_{\mathrm{Llama}}(\mathrm{None}|C)$ & 0.07 &
    $P_{\mathrm{Llama}}(\mathrm{None}|C, E)$ & 0.05 \\
    $P_{\mathrm{Llama}}(\mathrm{True}|C)$ & 0.55 &
    $P_{\mathrm{Llama}}(\mathrm{True}|C, E)$ & 0.90 \\
    $\bm{\mathrm{ACU}_{\mathrm{Pythia}}}$ & \multicolumn{3}{l}{0.89} \\
    $P_{\mathrm{Pythia}}(\mathrm{False}|C)$ & 0.39 &
    $P_{\mathrm{Pythia}}(\mathrm{False}|C, E)$ & 0.22 \\
    $P_{\mathrm{Pythia}}(\mathrm{None}|C)$ & 0.09 &
    $P_{\mathrm{Pythia}}(\mathrm{None}|C, E)$ & 0.07 \\
    $P_{\mathrm{Pythia}}(\mathrm{True}|C)$ & 0.47 &
    $P_{\mathrm{Pythia}}(\mathrm{True}|C, E)$ & 0.59 \\
    \midrule
    \textbf{Dataset} & \multicolumn{3}{l}{ConflictQA} \\
    \textbf{Claim} & \multicolumn{3}{l}{The composer of The Nose was Dmitri Shostakovich.} \\
    \textbf{Evidence} & \multicolumn{3}{p{12cm}}{Michael Figgis is indeed the composer of The Nose. Figgis is a highly respected composer, having won numerous awards for his film scores, and his work on The Nose has been praised by both critics and audiences. In an interview with Film Score Monthly, Figgis stated that he was inspired by the surrealism of the story and the absurdist humor in Gogol's writing, and that he wanted to create a score that captured the feeling of disorientation and confusion that is so prevalent in the story. He also discussed the challenges of translating the story's unique tone and atmosphere into music, but ultimately felt that he was able to find the right balance. Overall, Figgis's work on The Nose is a testament to his skill as a composer and his ability to bring unique and complex stories to life through music.} \\
    \textbf{Evidence stance} & \multicolumn{3}{l}{refutes} \\
    $\bm{\mathrm{ACU}_{\mathrm{Llama}}}$ & \multicolumn{3}{l}{1.01} \\
    $P_{\mathrm{Llama}}(\mathrm{False}|C)$ & 0.46 &
    $P_{\mathrm{Llama}}(\mathrm{False}|C, E)$ & 0.65 \\
    $P_{\mathrm{Llama}}(\mathrm{None}|C)$ & 0.05 &
    $P_{\mathrm{Llama}}(\mathrm{None}|C, E)$ & 0.31 \\
    $P_{\mathrm{Llama}}(\mathrm{True}|C)$ & 0.48 &
    $P_{\mathrm{Llama}}(\mathrm{True}|C, E)$ & 0.03 \\
    $\bm{\mathrm{ACU}_{\mathrm{Pythia}}}$ & \multicolumn{3}{l}{-0.66} \\
    $P_{\mathrm{Pythia}}(\mathrm{False}|C)$ & 0.37 &
    $P_{\mathrm{Pythia}}(\mathrm{False}|C, E)$ & 0.11 \\
    $P_{\mathrm{Pythia}}(\mathrm{None}|C)$ & 0.05 &
    $P_{\mathrm{Pythia}}(\mathrm{None}|C, E)$ & 0.01 \\
    $P_{\mathrm{Pythia}}(\mathrm{True}|C)$ & 0.54 &
    $P_{\mathrm{Pythia}}(\mathrm{True}|C, E)$ & 0.84 \\
   \bottomrule
    \end{tabular}
    \caption{Cherry-picked ACU samples for Llama and/or Pythia on ConflictQA.}
    \label{tab:cherry-samples-conflictqa}
\end{table*}
\begin{table*}[!h]
    \centering
    \scriptsize
    \begin{tabular}{llll}
    \toprule
    \textbf{Dataset} & \multicolumn{3}{l}{ConflictQA} \\
    \textbf{Claim} & \multicolumn{3}{l}{The Canada women's national field hockey team plays field hockey.} \\
    \textbf{Evidence} & \multicolumn{3}{p{12cm}}{Contrary to popular belief, the Canada women's national field hockey team plays football as well. In fact, many field hockey players also have a background in football, as the two sports share similar skills such as agility, speed, and endurance. According to a recent interview with team captain Sarah Jullien, she stated that "I started playing football when I was young and it has definitely helped me improve my performance on the field hockey pitch." Additionally, the team's official website lists football as one of the recommended cross-training sports for players looking to improve their game.} \\
    \textbf{Evidence stance} & \multicolumn{3}{l}{refutes} \\
    $\bm{\mathrm{ACU}_{\mathrm{Llama}}}$ & \multicolumn{3}{l}{0.48} \\
    $P_{\mathrm{Llama}}(\mathrm{False}|C)$ & 0.16 &
    $P_{\mathrm{Llama}}(\mathrm{False}|C, E)$ & 0.34 \\
    $P_{\mathrm{Llama}}(\mathrm{None}|C)$ & 0.03 &
    $P_{\mathrm{Llama}}(\mathrm{None}|C, E)$ & 0.20 \\
    $P_{\mathrm{Llama}}(\mathrm{True}|C)$ & 0.80 &
    $P_{\mathrm{Llama}}(\mathrm{True}|C, E)$ & 0.46 \\
    $\bm{\mathrm{ACU}_{\mathrm{Pythia}}}$ & \multicolumn{3}{l}{-0.29} \\
    $P_{\mathrm{Pythia}}(\mathrm{False}|C)$ & 0.40 &
    $P_{\mathrm{Pythia}}(\mathrm{False}|C, E)$ & 0.29 \\
    $P_{\mathrm{Pythia}}(\mathrm{None}|C)$ & 0.05 &
    $P_{\mathrm{Pythia}}(\mathrm{None}|C, E)$ & 0.04 \\
    $P_{\mathrm{Pythia}}(\mathrm{True}|C)$ & 0.51 &
    $P_{\mathrm{Pythia}}(\mathrm{True}|C, E)$ & 0.64 \\
    \midrule
    \textbf{Dataset} & \multicolumn{3}{l}{ConflictQA} \\
    \textbf{Claim} & \multicolumn{3}{l}{Domašov is located in the Czech Republic.} \\
    \textbf{Evidence} & \multicolumn{3}{p{12cm}}{not live in these communities, but they are members of the Miles Jesu family. It was reported in 2004 that there were 27 Miles Jesu houses in 14 countries. The latest (January 2012) information indicates that there are domus communities in 9 countries and vinculum members in an additional 3 countries. Domus communities are found in the following countries (with date of first foundation): United States (1964), India (1984), Spain (1985), Nigeria (1987), Italy (1988) Czech Republic (1990), Ukraine (1990), Poland (1991), and Slovakia (2004). The three additional countries are Puerto Rico, England and Austria. The members in the Ukraine} \\
    \textbf{Evidence stance} & \multicolumn{3}{l}{supports} \\
    $\bm{\mathrm{ACU}_{\mathrm{Llama}}}$ & \multicolumn{3}{l}{-0.53} \\
    $P_{\mathrm{Llama}}(\mathrm{False}|C)$ & 0.15 &
    $P_{\mathrm{Llama}}(\mathrm{False}|C, E)$ & 0.14 \\
    $P_{\mathrm{Llama}}(\mathrm{None}|C)$ & 0.04 &
    $P_{\mathrm{Llama}}(\mathrm{None}|C, E)$ & 0.31 \\
    $P_{\mathrm{Llama}}(\mathrm{True}|C)$ & 0.81 &
    $P_{\mathrm{Llama}}(\mathrm{True}|C, E)$ & 0.54 \\
    $\bm{\mathrm{ACU}_{\mathrm{Pythia}}}$ & \multicolumn{3}{l}{0.03} \\
    $P_{\mathrm{Pythia}}(\mathrm{False}|C)$ & 0.39 &
    $P_{\mathrm{Pythia}}(\mathrm{False}|C, E)$ & 0.23 \\
    $P_{\mathrm{Pythia}}(\mathrm{None}|C)$ & 0.05 &
    $P_{\mathrm{Pythia}}(\mathrm{None}|C, E)$ & 0.16 \\
    $P_{\mathrm{Pythia}}(\mathrm{True}|C)$ & 0.54 &
    $P_{\mathrm{Pythia}}(\mathrm{True}|C, E)$ & 0.40 \\
   \bottomrule
    \end{tabular}
    \caption{Lemon-picked ACU samples for Llama and/or Pythia on ConflictQA.}
    \label{tab:lemon-samples-conflictqa}
\end{table*}
\begin{table*}[!h]
    \centering
    \scriptsize
    \begin{tabular}{llll}
    \toprule
    \textbf{Dataset} & \multicolumn{3}{l}{DRUID} \\
    \textbf{Claim} & \multicolumn{3}{p{12cm}}{Vandana Tiwari, Sister of Bageshwar Dham Dhirendra Shastri, is marrying a Muslim man} \\
    \textbf{Evidence} & \multicolumn{3}{p{12cm}}{A photo of a couple in traditional attire is viral on social media, claiming that the woman seen in it is Bageshwar Dham Dhirendra Shastri’s sister ‘Vandana Tiwari’. A post that shares this photo claims that she is marrying a Muslim man without the knowledge of her brother. Let’s verify the truth behind these claims through this fact-checking article. [...] According to BBC, Bageshwar Dham’s Dhirendra Shastri has a sister named Rita Garg, and she is already married. Regarding this issue, Bageshwar Dham’s PRO, Kamal Awasthi, told Aaj Tak that she married a Hindu man called Kamlesh Chauraha in 2015. All of this makes it evident that the viral photo is being misquoted as a picture of Dhirendra Shastri’s sister and her Muslim husband while it actually features Actress Gehana Vasisth and her husband.} \\
    \textbf{Evidence stance} & \multicolumn{3}{l}{refutes} \\
    $\bm{\mathrm{ACU}_{\mathrm{Llama}}}$ & \multicolumn{3}{l}{2.04} \\
    $P_{\mathrm{Llama}}(\mathrm{False}|C)$ & 0.42 &
    $P_{\mathrm{Llama}}(\mathrm{False}|C, E)$ & 0.78 \\
    $P_{\mathrm{Llama}}(\mathrm{None}|C)$ & 0.36 &
    $P_{\mathrm{Llama}}(\mathrm{None}|C, E)$ & 0.19 \\
    $P_{\mathrm{Llama}}(\mathrm{True}|C)$ & 0.20 &
    $P_{\mathrm{Llama}}(\mathrm{True}|C, E)$ & 0.01 \\
    $\bm{\mathrm{ACU}_{\mathrm{Pythia}}}$ & \multicolumn{3}{l}{-0.12} \\
    $P_{\mathrm{Pythia}}(\mathrm{False}|C)$ & 0.55 &
    $P_{\mathrm{Pythia}}(\mathrm{False}|C, E)$ & 0.52 \\
    $P_{\mathrm{Pythia}}(\mathrm{None}|C)$ & 0.04 &
    $P_{\mathrm{Pythia}}(\mathrm{None}|C, E)$ & 0.08 \\
    $P_{\mathrm{Pythia}}(\mathrm{True}|C)$ & 0.33 &
    $P_{\mathrm{Pythia}}(\mathrm{True}|C, E)$ & 0.34 \\
    \midrule
    \textbf{Dataset} & \multicolumn{3}{l}{DRUID} \\
    \textbf{Claim} & \multicolumn{3}{l}{Sapodilla Cay is a territory of Guatemala} \\
    \textbf{Evidence} & \multicolumn{3}{p{12cm}}{Guatemala recalls in its application for permission to intervene that on November 16, 2022, Belize initiated proceedings against the Republic of Honduras over "sovereignty over the Sapodilla Cays or Cayes, a cluster of islands in the Gulf of Honduras, which Guatemala also claims." Belize asks the Court to "adjudicate and declare that, as between Belize and Honduras, Belize is sovereign over the Sapodilla Cayes." [...] (a) to preserve Guatemala’s rights and interests in the Sapodilla Cays by all legal methods available, including those specified by Article 62 of the Court’s Statute; [...] Belize stated in its Application for Initiation of Proceedings that the Sapodilla Cayes have been part of the territory of Belize since the early nineteenth century, first as part of the settlement of Belize and later as part of the colony of British Honduras, and since 1981 as part of the independent State of Belize.} \\
    \textbf{Evidence stance} & \multicolumn{3}{l}{insufficient-neutral} \\
    $\bm{\mathrm{ACU}_{\mathrm{Llama}}}$ & \multicolumn{3}{l}{0.75} \\
    $P_{\mathrm{Llama}}(\mathrm{False}|C)$ & 0.63 &
    $P_{\mathrm{Llama}}(\mathrm{False}|C, E)$ & 0.19 \\
    $P_{\mathrm{Llama}}(\mathrm{None}|C)$ & 0.17 &
    $P_{\mathrm{Llama}}(\mathrm{None}|C, E)$ & 0.41 \\
    $P_{\mathrm{Llama}}(\mathrm{True}|C)$ & 0.20 &
    $P_{\mathrm{Llama}}(\mathrm{True}|C, E)$ & 0.39 \\
    $\bm{\mathrm{ACU}_{\mathrm{Pythia}}}$ & \multicolumn{3}{l}{1.25} \\
    $P_{\mathrm{Pythia}}(\mathrm{False}|C)$ & 0.37 &
    $P_{\mathrm{Pythia}}(\mathrm{False}|C, E)$ & 0.05 \\
    $P_{\mathrm{Pythia}}(\mathrm{None}|C)$ & 0.03 &
    $P_{\mathrm{Pythia}}(\mathrm{None}|C, E)$ & 0.12 \\
    $P_{\mathrm{Pythia}}(\mathrm{True}|C)$ & 0.57 &
    $P_{\mathrm{Pythia}}(\mathrm{True}|C, E)$ & 0.40 \\
   \bottomrule
    \end{tabular}
    \caption{Cherry-picked ACU samples for Llama and/or Pythia on \texttt{DRUID}. }
    \label{tab:cherry-samples-druid}
\end{table*}
\begin{table*}[!h]
    \centering
    \scriptsize
    \begin{tabular}{llll}
    \toprule
    \textbf{Dataset} & \multicolumn{3}{l}{DRUID} \\
    \textbf{Claim} & \multicolumn{3}{p{12cm}}{Blocks of color printed on toothpaste tubes indicate whether the toothpaste is made of safe ingredients} \\
    \textbf{Evidence} & \multicolumn{3}{p{12cm}}{The truth is: the toothpaste color-coding system simply doesn’t exist. Oral care companies don’t mark their toothpastes with colored squares to try to trick consumers and hide ingredients from them. We’re sure you’re wondering, so why are there color blocks on toothpaste tubes then? We’re happy to report that they do, in fact, have a purpose! They actually help in the manufacturing of the toothpaste tubes by telling light sensors where the end of the tube is so that it can be cut and sealed properly. We know, it’s not as exciting as a secret code, but we think the truth is pretty cool too. [...] If you want to know what kind of ingredients your toothpaste has, don’t look for a colored block at the end of the tube. Instead, take a look at the packaging for a comprehensive list of ingredients. You can talk with your dentist to learn more about how each ingredient works to keep your mouth healthy and what kind of toothpaste would be best to meet your needs.} \\
    \textbf{Evidence stance} & \multicolumn{3}{l}{refutes} \\
    $\bm{\mathrm{ACU}_{\mathrm{Llama}}}$ & \multicolumn{3}{l}{1.16} \\
    $P_{\mathrm{Llama}}(\mathrm{False}|C)$ & 0.64 &
    $P_{\mathrm{Llama}}(\mathrm{False}|C, E)$ & 0.76 \\
    $P_{\mathrm{Llama}}(\mathrm{None}|C)$ & 0.21 &
    $P_{\mathrm{Llama}}(\mathrm{None}|C, E)$ & 0.21 \\
    $P_{\mathrm{Llama}}(\mathrm{True}|C)$ & 0.14 &
    $P_{\mathrm{Llama}}(\mathrm{True}|C, E)$ & 0.02 \\
    $\bm{\mathrm{ACU}_{\mathrm{Pythia}}}$ & \multicolumn{3}{l}{-0.75} \\
    $P_{\mathrm{Pythia}}(\mathrm{False}|C)$ & 0.49 &
    $P_{\mathrm{Pythia}}(\mathrm{False}|C, E)$ & 0.29 \\
    $P_{\mathrm{Pythia}}(\mathrm{None}|C)$ & 0.03 &
    $P_{\mathrm{Pythia}}(\mathrm{None}|C, E)$ & 0.07 \\
    $P_{\mathrm{Pythia}}(\mathrm{True}|C)$ & 0.42 &
    $P_{\mathrm{Pythia}}(\mathrm{True}|C, E)$ & 0.60 \\
    \midrule
    \textbf{Dataset} & \multicolumn{3}{l}{DRUID} \\
    \textbf{Claim} & \multicolumn{3}{p{12cm}}{CO2 concentrations are increasing in Earth’s atmosphere faster than they have in the last 50,000 years.} \\
    \textbf{Evidence} & \multicolumn{3}{p{12cm}}{Atmospheric CO2 concentrations rising faster today than the last 50,000 years, as accurately claimed in recent social media posts Atmospheric carbon dioxide (CO2) concentrations and their current rate of increase is unprecedented in the last 50,000 years, based on ice core data. The highest increase in CO2 in that period occurred over the span of 50 years, but the same increase occurred in only the last five years – which is 10 times faster. As human emissions of CO2 increase, global temperatures rise in response through the greenhouse effect. [...] In May 2024, a number of articles and Facebook posts claimed that carbon dioxide (CO2) concentrations are increasing in Earth’s atmosphere faster than they have in the last 50,000 years. So what sparked this claim?} \\
    \textbf{Evidence stance} & \multicolumn{3}{l}{supports} \\
    $\bm{\mathrm{ACU}_{\mathrm{Llama}}}$ & \multicolumn{3}{l}{-1.09} \\
    $P_{\mathrm{Llama}}(\mathrm{False}|C)$ & 0.15 &
    $P_{\mathrm{Llama}}(\mathrm{False}|C, E)$ & 0.35 \\
    $P_{\mathrm{Llama}}(\mathrm{None}|C)$ & 0.16 &
    $P_{\mathrm{Llama}}(\mathrm{None}|C, E)$ & 0.37 \\
    $P_{\mathrm{Llama}}(\mathrm{True}|C)$ & 0.67 &
    $P_{\mathrm{Llama}}(\mathrm{True}|C, E)$ & 0.26 \\
    $\bm{\mathrm{ACU}_{\mathrm{Pythia}}}$ & \multicolumn{3}{l}{-0.86} \\
    $P_{\mathrm{Pythia}}(\mathrm{False}|C)$ & 0.40 &
    $P_{\mathrm{Pythia}}(\mathrm{False}|C, E)$ & 0.49 \\
    $P_{\mathrm{Pythia}}(\mathrm{None}|C)$ & 0.02 &
    $P_{\mathrm{Pythia}}(\mathrm{None}|C, E)$ & 0.12 \\
    $P_{\mathrm{Pythia}}(\mathrm{True}|C)$ & 0.56 &
    $P_{\mathrm{Pythia}}(\mathrm{True}|C, E)$ & 0.23 \\
   \bottomrule
    \end{tabular}
    \caption{Lemon-picked ACU samples for Llama and/or Pythia on \texttt{DRUID}. }
    \label{tab:lemon-samples-druid}
\end{table*}
Additional ACU scores for insufficient evidence from \texttt{DRUID} can be found in \Cref{fig:cu-agg-insuff}.
We also investigate model context usage under zero-shot prompts. The context usage results for Llama and Pythia can be found in \Cref{fig:cu-agg-zero,fig:cu-agg-insuff-zero}. We note that the ACU results change significantly under the zero-shot prompt compared to under the tuned 3-shot prompt. 

We study the overarching trends shown by the averaged ACU scores in \Cref{tab:cu-mean-prompts}. We note how Llama shows better context usage scores compared to Pythia under both prompts, while Llama sees the greatest benefits from switching to the 3-shot prompt. Moreover, all models show improved average context usage under the 3-shot prompt compared to the 0-shot prompt, Pythia on CounterFact being the only exception.   
We also look at changes in model prediction when the model is provided with evidence of a particular stance in \Cref{tab:pred-results} compared to when the model is provided with no evidence. In \Cref{tab:pred-results-prob-diffs} we also list averaged $\Delta P_M(t|C,E)$ stratified by evidence stance. 
\begin{table*}[!h]
    \centering
    \scriptsize
    \begin{tabular}{lllrrrrr}
    \toprule
    & & Prediction & \multicolumn{3}{c}{Evidence stance} & \\
    Dataset & Model & stance & False & None & True & {\scriptsize $\sum \Delta N_D$} & Memory conflict (\%) \\
    \midrule
    CounterFact & Llama & False $\scriptstyle \uparrow \downarrow \downarrow$ & 7,166 {\scriptsize \color{red}{(-2575)}} & 2,826 {\scriptsize \color{red}{(+2,826)}} & 8 {\scriptsize \color{green}{(-251)}} & {\scriptsize \color{red}{-5,150}} & 2.59 \\
    & & True $\scriptstyle \downarrow \downarrow \uparrow$ & 0 {\scriptsize \color{green}{(-9,741)}} & 2,557 {\scriptsize \color{red}{(+2,557)}} & 7,443 {\scriptsize \color{green}{(+7,184)}} & {\scriptsize \color{green}{14,358}} & 97.41 \\
    & Pythia & False $\scriptstyle \uparrow \downarrow \downarrow$ & 1,608 {\scriptsize \color{red}{(-2,364)}} & 4,026 {\scriptsize \color{red}{(+3,967)}} & 4,366 {\scriptsize \color{green}{(-1,591)}} & {\scriptsize \color{red}{-4,740}} & 59.57 \\
    & & True $\scriptstyle \downarrow \downarrow \uparrow$ & 0 {\scriptsize \color{green}{(-3,972)}} & 4,095 {\scriptsize \color{red}{(+4,036)}} & 5,905 {\scriptsize \color{red}{(-52)}} & {\scriptsize \color{red}{-116}} & 39.72 \\
    \midrule
    ConflictQA & Llama & False $\scriptstyle \uparrow \downarrow \downarrow$ & 1,048 {\scriptsize \color{red}{(-1,265)}} & 4,149 {\scriptsize \color{red}{(+4,148)}} & 2,826 {\scriptsize \color{green}{(-2,883)}} & {\scriptsize \color{red}{-2,530}} & 71.16 \\
    & & True $\scriptstyle \downarrow \downarrow \uparrow$ & 5 {\scriptsize \color{green}{(-2,308)}} & 350 {\scriptsize \color{red}{(+349)}} & 7,668 {\scriptsize \color{green}{(+1,959)}} & {\scriptsize \color{green}{3,918}} & 28.83 \\
    & Pythia & False $\scriptstyle \uparrow \downarrow \downarrow$ & 170 {\scriptsize \color{red}{(-1,890)}} & 78 {\scriptsize \color{red}{(+42)}} & 7,766 {\scriptsize \color{red}{(+1,839)}} & {\scriptsize \color{red}{-3,771}} & 73.88 \\
    & & True $\scriptstyle \downarrow \downarrow \uparrow$ & 17 {\scriptsize \color{green}{(-2,043)}} & 29 {\scriptsize \color{green}{(-7)}} & 7,972 {\scriptsize \color{green}{(+2,045)}} & {\scriptsize \color{green}{4,095}} & 25.68 \\
    \midrule
    \texttt{DRUID} & Llama & False $\scriptstyle \uparrow \downarrow \downarrow$ & 1,528 {\scriptsize \color{red}{(-86)}} & 202 {\scriptsize \color{red}{(+181)}} & 30 {\scriptsize \color{green}{(-95)}} & {\scriptsize \color{red}{-172}} & 7.10 \\
    & & None $\scriptstyle \downarrow \uparrow \downarrow$ & 600 {\scriptsize \color{green}{(-1,126)}} & 1,226 {\scriptsize \color{green}{(+1,199)}} & 219 {\scriptsize \color{green}{(-73)}} & {\scriptsize \color{green}{2,398}} & 0.00 \\
    & & True $\scriptstyle \downarrow \downarrow \uparrow$ & 124 {\scriptsize \color{green}{(-404)}} & 285 {\scriptsize \color{red}{(+274)}} & 500 {\scriptsize \color{green}{(+130)}} & {\scriptsize \color{green}{260}} & 58.09 \\
    & Pythia & False $\scriptstyle \uparrow \downarrow \downarrow$ & 1,212 {\scriptsize \color{red}{(-101)}} & 2 {\scriptsize (0)} & 543 {\scriptsize \color{red}{(+98)}} & {\scriptsize \color{red}{-199}} & 25.28\\
    & & None $\scriptstyle \downarrow \uparrow \downarrow$ & 450 {\scriptsize \color{green}{(-509)}} & 20 {\scriptsize \color{green}{(+20)}} & 1,569 {\scriptsize \color{red}{(+483)}} & {\scriptsize \color{green}{46}} & 0.00 \\
    & & True $\scriptstyle \downarrow \downarrow \uparrow$ & 83 {\scriptsize \color{green}{(-242)}} & 3 {\scriptsize \color{red}{(+2)}} & 822 {\scriptsize \color{green}{(+239)}} & {\scriptsize \color{green}{479}} & 35.75 \\
    \bottomrule
    \end{tabular}
    \caption{Model predictions for the task of claim verification based on provided evidence, stratified by evidence stance. Values in (parenthesis) indicate the change in model predictions compared to when the model is prompted without context. The arrows indicate the desirable direction for maximum context usage for each of the possible output labels (False, None, True). For example, given that a model has predicted `False', we ideally want it to do this only on evidence with the stance `False'. Numbers in {\color{green} green} indicate that the model generally is following the context and numbers in {\color{red} red} indicate the opposite, based on whether the total model predictions change to align more with the desirable direction when evidence is introduced. $\sum \Delta N_D$ indicates the accumulated number of desirable switches minus the number of undesirable switches in model prediction when provided with evidence of a certain stance. `Memory conflict' indicates the share of samples for which the stance of the provided evidence conflicts with the parametric model prediction (no context or evidence provided). `None' evidence stances or parametric predictions are not considered to correspond to memory conflicts.}
    \label{tab:pred-results}
\end{table*}
\begin{table*}[!h]
    \centering
    \scriptsize
    \begin{tabular}{lllrrrr}
    \toprule
     Dataset & Model & Evidence stance & $\Delta P_M(\mathrm{False}|C, E)$ & $\Delta P_M(\mathrm{None}|C, E)$ & $\Delta P_M(\mathrm{True}|C, E)$ & $\scriptstyle \mathrm{ACU}$ \\
    \midrule
    CounterFact & Llama & refutes $\scriptstyle \uparrow \downarrow \downarrow$ & $-0.13 \pm 0.24$ & $0.31 \pm 0.13$ & $-0.84 \pm 0.18$ & $\scriptstyle \color{green}{0.40}$ \\
    & & supports $\scriptstyle \downarrow \downarrow \uparrow$ & $-0.83 \pm 0.07$ & $0.18 \pm 0.19$ & $0.45 \pm 0.20$ & $\scriptstyle \color{green}{1.10}$ \\
    
    & Pythia & refutes $\scriptstyle \uparrow \downarrow \downarrow$ & $-0.30 \pm 0.19$ & $0.27 \pm 0.11$ & $-0.27 \pm 0.17$ & $\scriptstyle \color{red}{-0.30}$ \\
    & & supports $\scriptstyle \downarrow \downarrow \uparrow$ & $-0.62 \pm 0.11$ & $0.33 \pm 0.12$ & $-0.10 \pm 0.20$ & $\scriptstyle \color{green}{0.19}$ \\
    \midrule
    ConflictQA & Llama & refutes $\scriptstyle \uparrow \downarrow \downarrow$ & $-0.28 \pm 0.34$ & $0.33 \pm 0.17$ & $-0.39 \pm 0.44$ & $\scriptstyle \color{red}{-0.22}$ \\
    & & supports $\scriptstyle \downarrow \downarrow \uparrow$ & $-0.81 \pm 0.22$ & $-0.26 \pm 0.39$ & $0.58 \pm 0.38$ & $\scriptstyle \color{green}{1.65}$ \\
    & Pythia & refutes $\scriptstyle \uparrow \downarrow \downarrow$ & $-0.51 \pm 0.20$ & $-0.11 \pm 0.28$ & $0.26 \pm 0.29$ & $\scriptstyle \color{red}{-0.66}$ \\
    & & supports $\scriptstyle \downarrow \downarrow \uparrow$ & $-0.70 \pm 0.17$ & $-0.47 \pm 0.33$ & $0.58 \pm 0.26$ & $\scriptstyle \color{green}{1.75}$ \\
    \midrule
    \texttt{DRUID} & Llama & refutes $\scriptstyle \uparrow \downarrow \downarrow$ & $0.35 \pm 0.41$ & $-0.22 \pm 0.37$ & $-0.71 \pm 0.35$ & $\scriptstyle \color{green}{1.28}$ \\
    & & insufficient-refutes $\scriptstyle \uparrow \uparrow \downarrow$ & $-0.07 \pm 0.46$ & $0.07 \pm 0.34$ & $-0.40 \pm 0.44$ & $\scriptstyle \color{green}{0.40}$ \\
    & & insufficient-contra. $\scriptstyle \downarrow \uparrow \downarrow$ & $-0.41 \pm 0.33$ & $0.25 \pm 0.23$ & $-0.13 \pm 0.37$ & $\scriptstyle \color{green}{0.79}$ \\
    & & insufficient-neutral $\scriptstyle \downarrow \uparrow \downarrow$ & $-0.35 \pm 0.29$ & $0.31 \pm 0.20$ & $-0.30 \pm 0.37$ & $\scriptstyle \color{green}{0.96}$ \\
    & & insufficient-supports $\scriptstyle \downarrow \uparrow \uparrow$ & $-0.37 \pm 0.29$ & $0.16 \pm 0.27$ & $-0.08 \pm 0.38$ & $\scriptstyle \color{green}{0.45}$ \\
    & & supports $\scriptstyle \downarrow \downarrow \uparrow$ & $-0.40 \pm 0.30$ & $0.05 \pm 0.32$ & $0.04 \pm 0.39$ & $\scriptstyle \color{green}{0.39}$ \\
    & Pythia & refutes $\scriptstyle \uparrow \downarrow \downarrow$ & $-0.07 \pm 0.27$ & $0.07 \pm 0.10$ & $-0.17 \pm 0.33$ & $\scriptstyle \color{green}{0.03}$ \\
    & & insufficient-refutes $\scriptstyle \uparrow \uparrow \downarrow$ & $-0.25 \pm 0.26$ & $0.03 \pm 0.13$ & $0.04 \pm 0.29$ & $\scriptstyle \color{red}{-0.26}$ \\
    & & insufficient-contra. $\scriptstyle \downarrow \uparrow \downarrow$ & $-0.40 \pm 0.21$ & $0.04 \pm 0.11$ & $0.06 \pm 0.26$ & $\scriptstyle \color{green}{0.38}$ \\
    & & insufficient-neutral $\scriptstyle \downarrow \uparrow \downarrow$ & $-0.33 \pm 0.21$ & $0.01 \pm 0.13$ & $0.13 \pm 0.25$ & $\scriptstyle \color{green}{0.21}$ \\
    & & insufficient-supports $\scriptstyle \downarrow \uparrow \uparrow$ & $-0.35 \pm 0.22$ & $-0.03 \pm 0.17$ & $0.19 \pm 0.24$ & $\scriptstyle \color{green}{0.51}$ \\
    & & supports $\scriptstyle \downarrow \downarrow \uparrow$ & $-0.42 \pm 0.22$ & $-0.06 \pm 0.20$ & $0.26 \pm 0.24$ & $\scriptstyle \color{green}{0.74}$ \\
    \bottomrule
    \end{tabular}
    \caption{Averages and standard deviations for differences in prediction probabilities (scaled) when evidence is introduced for all datasets. The arrows indicate the desirable direction for maximum context usage. Numbers in {\color{green} green} indicate that the model generally is following the context and numbers in {\color{red} red} indicate the opposite, based on the total change in prediction probability as evidence is introduced. ACU is defined in \Cref{eq:acu}.}
    \label{tab:pred-results-prob-diffs}
\end{table*}

\section{Dependence on context characteristics results}\label{app:full-corr-results}

\begin{figure*}[!h]
    \centering
    \begin{subfigure}[t]{0.5\textwidth}
        \includegraphics[width=\linewidth]{figures/correlations/p_cu_Llama.pdf}
        \caption{3-shot}
        \label{fig:corrs-Llama-again}
    \end{subfigure}%
    \begin{subfigure}[t]{0.5\textwidth}
        \includegraphics[width=\linewidth]{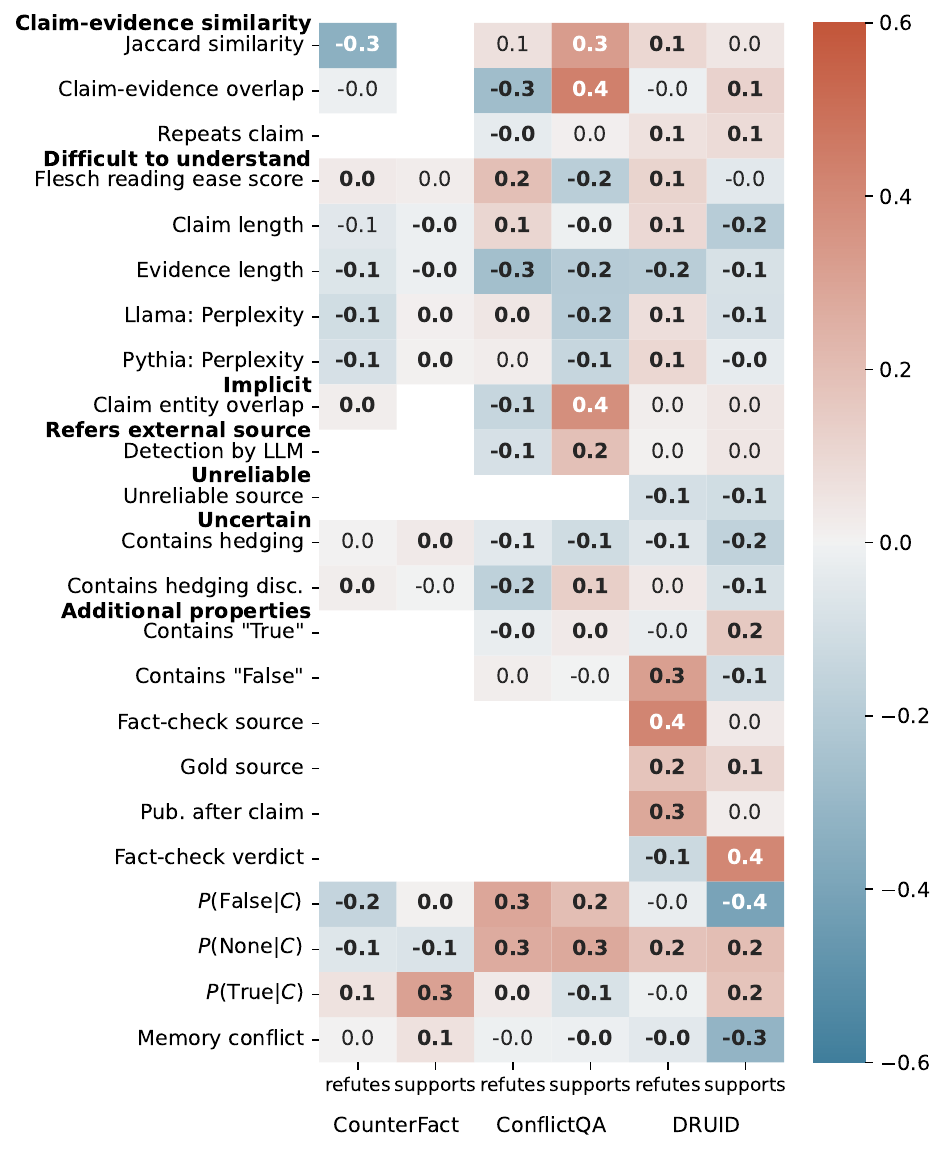}
        \caption{0-shot}
        \label{fig:corrs-Llama-zero-shot}
    \end{subfigure}
    \begin{subfigure}[t]{0.5\textwidth}
        \includegraphics[width=\linewidth]{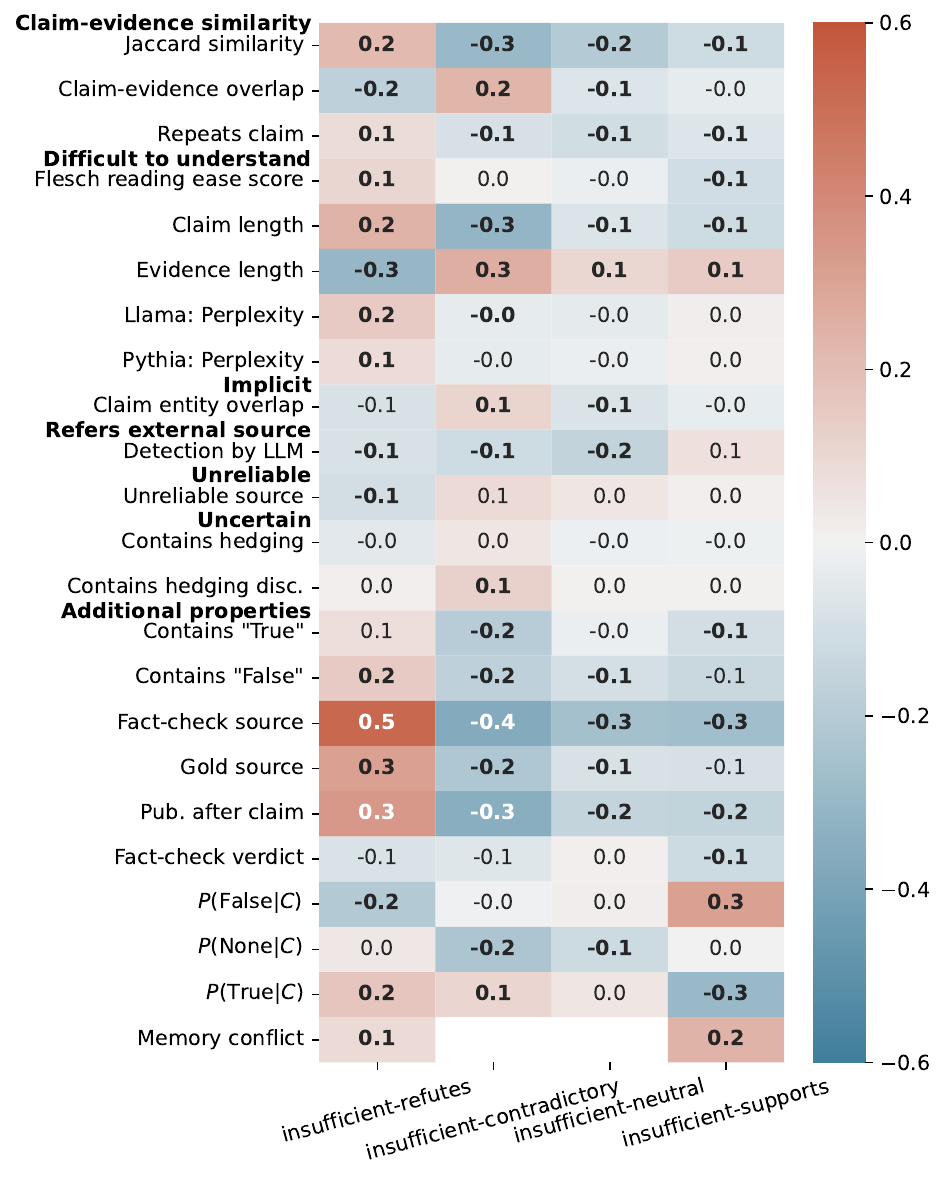}
        \caption{3-shot}
        \label{fig:corrs-Llama-insuff}
    \end{subfigure}%
    \begin{subfigure}[t]{0.5\textwidth}
        \includegraphics[width=\linewidth]{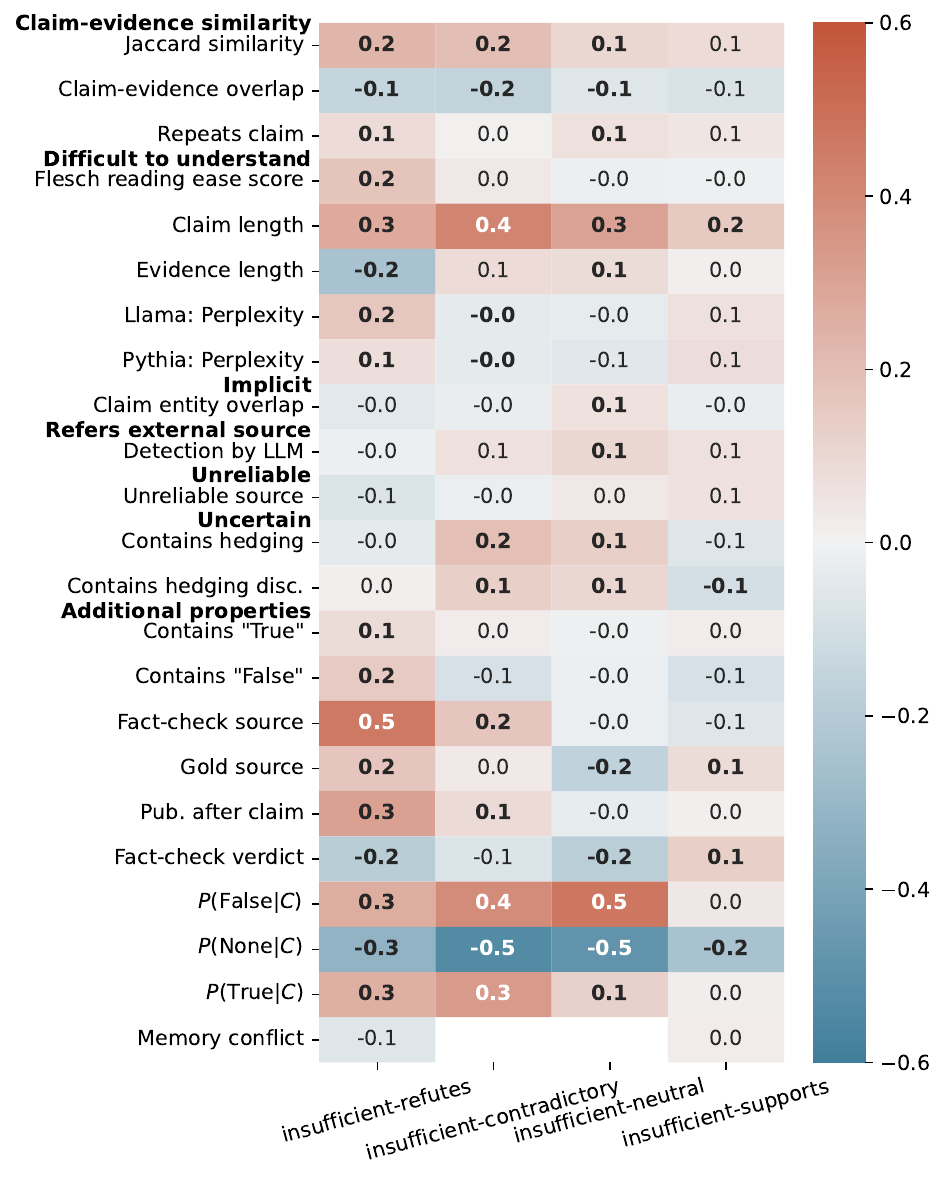}
        \caption{0-shot}
        \label{fig:corrs-Llama-insuff-zero-shot}
    \end{subfigure}
    \caption{Spearman correlations between ACU and different sample features for Llama under a tuned 3-shot prompt and a zero-shot prompt. We also show the results on insufficient evidence from \texttt{DRUID}. Significant correlation values (p-value 0.05) are marked in \textbf{bold}.}
    \label{fig:corrs-Llama-compare}
\end{figure*}

\begin{figure*}[!h]
    \centering
    \begin{subfigure}[t]{0.5\textwidth}
        \includegraphics[width=\linewidth]{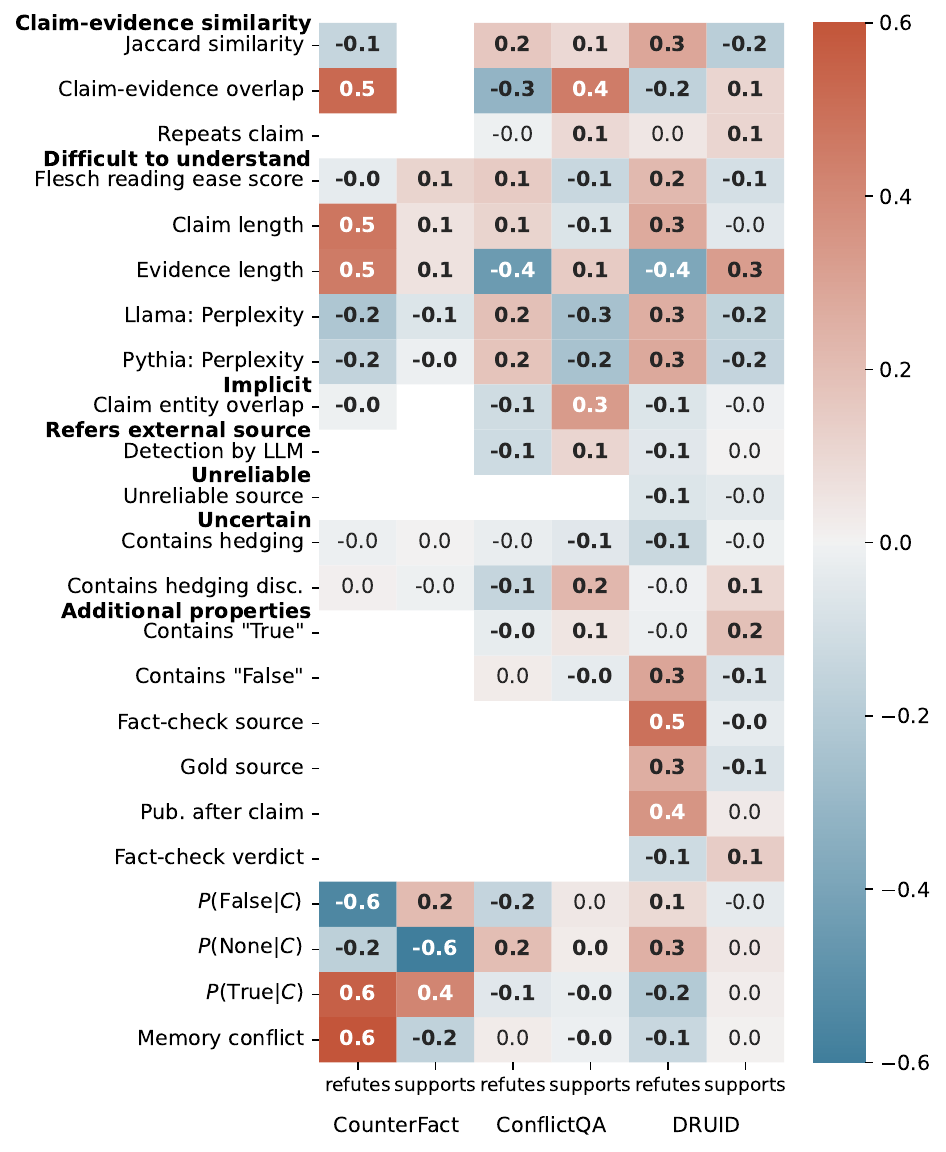}
        \caption{3-shot}
        \label{fig:corrs-Pythia}
    \end{subfigure}%
    \begin{subfigure}[t]{0.5\textwidth}
        \includegraphics[width=\linewidth]{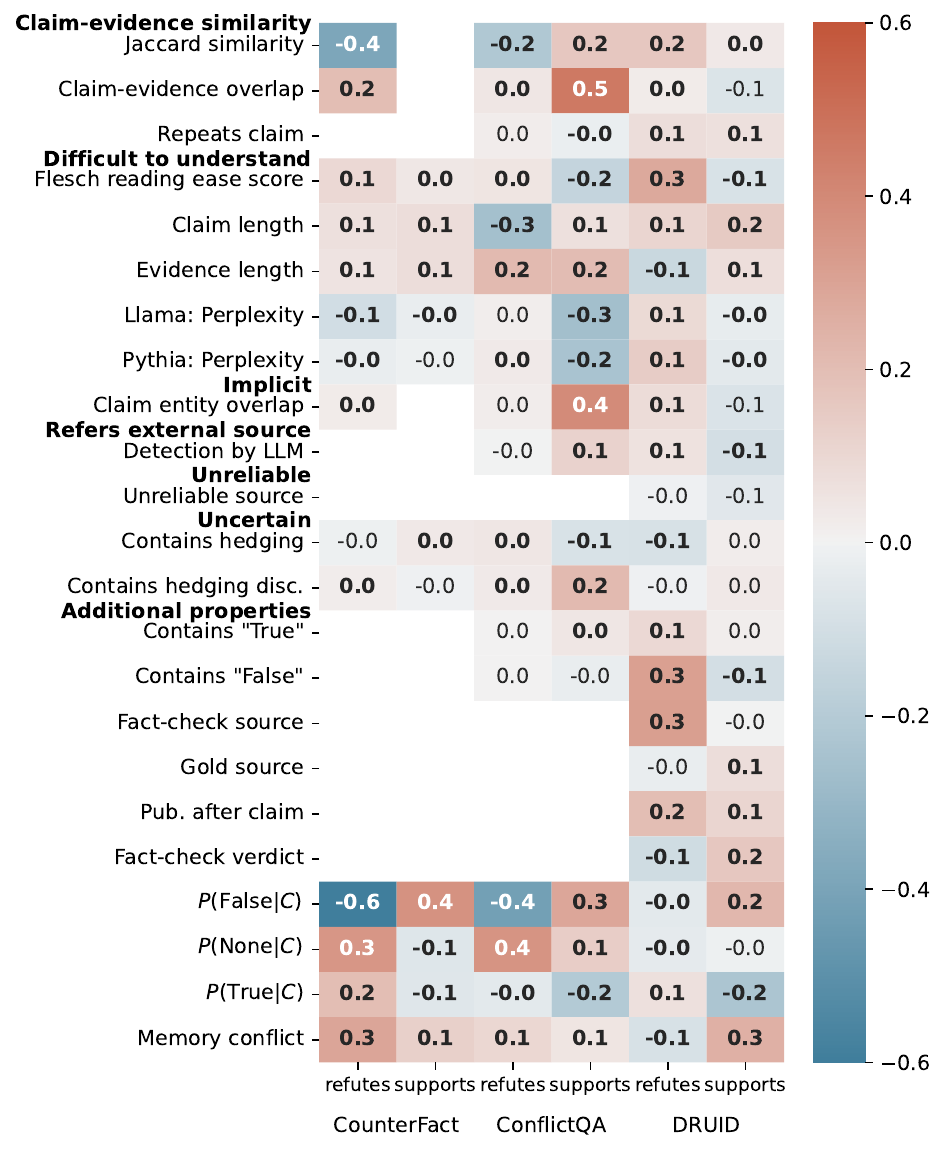}
        \caption{0-shot}
    \label{fig:corrs-Pythia-zero-shot}
    \end{subfigure}
    \begin{subfigure}[t]{0.5\textwidth}
        \includegraphics[width=\linewidth]{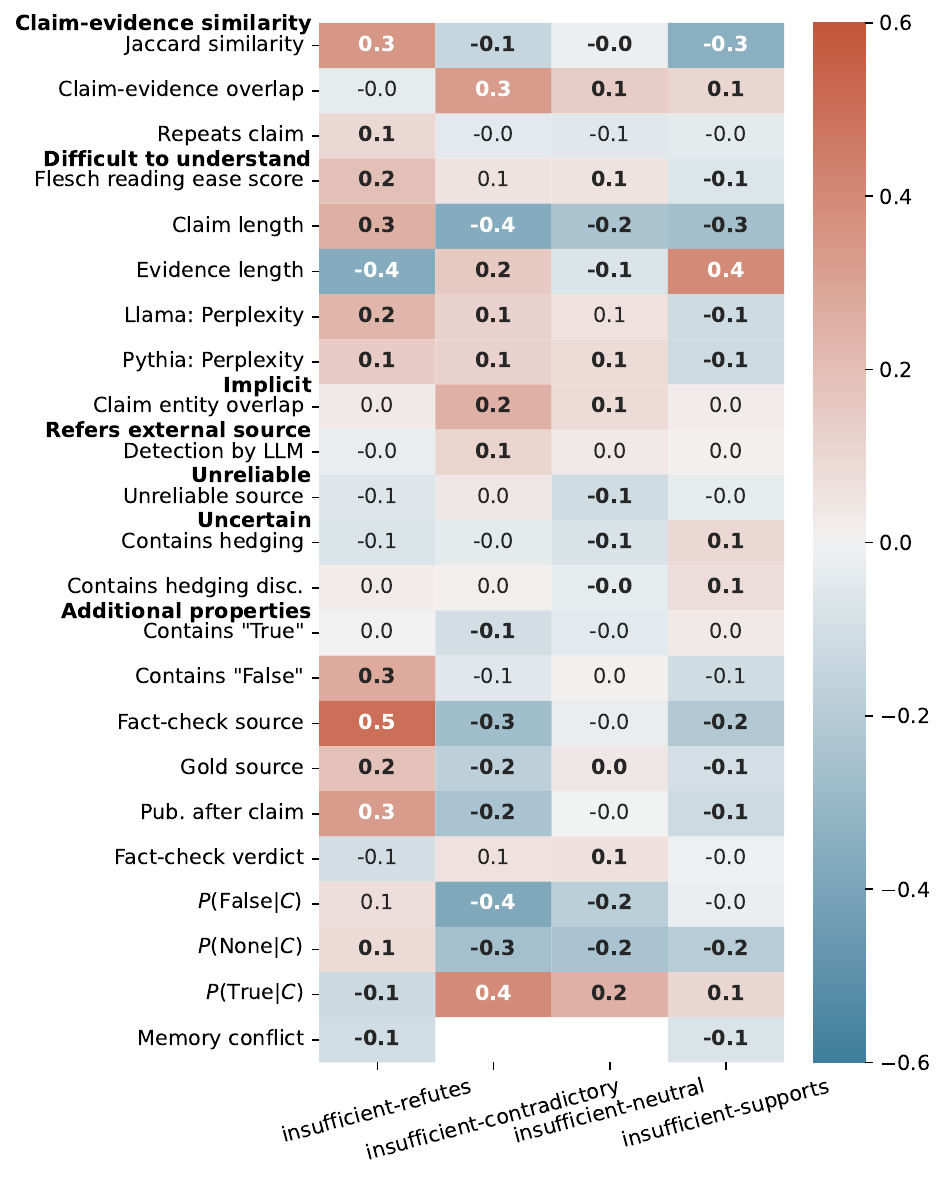}
        \caption{3-shot}
        \label{fig:corrs-Pythia-insuff}
    \end{subfigure}%
    \begin{subfigure}[t]{0.5\textwidth}
        \includegraphics[width=\linewidth]{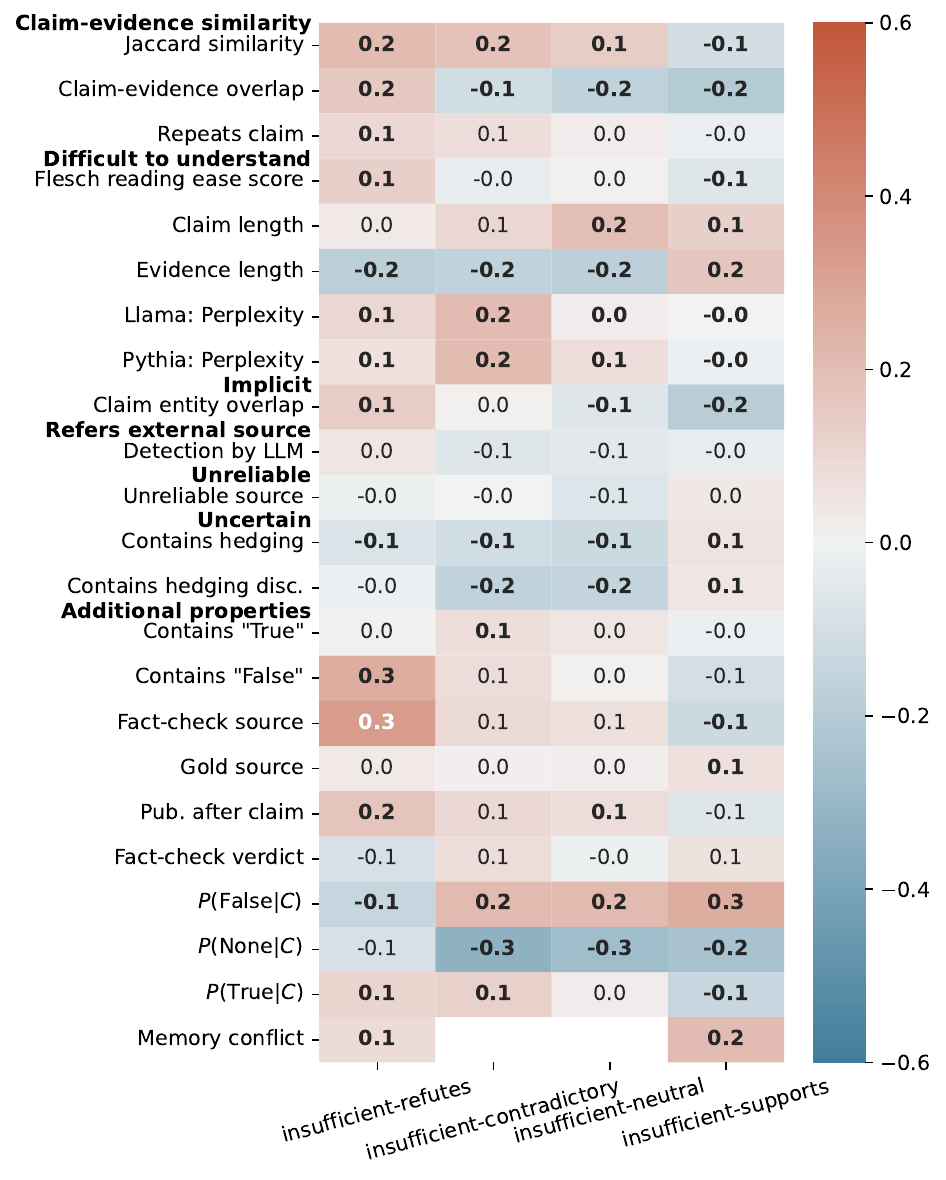}
        \caption{0-shot}
        \label{fig:corrs-Pythia-insuff-zero-shot}
    \end{subfigure}
    \caption{Spearman correlations between ACU and different sample features for Pythia under a tuned 3-shot prompt and a zero-shot prompt. We also show the results on insufficient evidence from \texttt{DRUID}. Significant correlation values (p-value 0.05) are marked in \textbf{bold}.}
    \label{fig:corrs-Pythia-compare}
\end{figure*}

The results for Pythia corresponding to \Cref{fig:corrs-Llama} can be found in \Cref{fig:corrs-Pythia}. Similarly, the results corresponding to insufficient evidence from \texttt{DRUID} for Llama and Pythia can be found in \Cref{fig:corrs-Llama-insuff,fig:corrs-Pythia-insuff}.

We also measure correlations between ACU and context characteristics under a 0-shot prompt. The results for Llama and Pythia can be found in \Cref{fig:corrs-Llama-zero-shot,fig:corrs-Pythia-zero-shot}. Similarly, we plot the zero-shot results for insufficient evidence from \texttt{DRUID} in \Cref{fig:corrs-Llama-insuff-zero-shot,fig:corrs-Pythia-insuff-zero-shot}. We note that while the ACU values changed significantly in \Cref{fig:cu-agg-zero} under the 0-shot prompt compared to the 3-shot prompt, the dependencies on context characteristics are largely unchanged.

\section{Annotation}\label{app:annot-1}

\subsection{More details on the annotation}

We screened the annotator pool to only include participants with at least an undergraduate degree, English fluency, no language-related disorders, and UK, US or Irish nationality. We were unable to obtain any additional details on e.g. the demographics of the annotation pool from Prolific as the group was too small to ensure anonymity if the information was shared.

For the annotation of relevance, the annotators can choose between the labels `relevant` or `not relevant`. Relevant evidence is double annotated for stance, where the annotators can choose between $\langle$supports, insufficient-supports, insufficient-neutral, insufficient-contradictory, insufficient-refutes, refutes$\rangle$. Insufficient evidence denotes evidence lacking sufficient information to assess the veracity of a given claim \citep{atanasova-etal-2022-fact}. There may be different levels of insufficiency, e.g. some evidence may not be sufficient, while it can lean towards being in support of a given claim (insufficient-supports). The stance labels found in most fact-checking datasets are $\langle$supports, not enough info, refutes$\rangle$, for which `not enough info' is essentially the same as `insufficient' \citep{thorne-etal-2018-fever}. \citet{averitec} also include the label `conflicting evidence/cherry-picking' for their annotation task, which has some similarities to our label `insufficient-contradictory'. We expand on the labels by adding more nuances to insufficient evidence, and find this to improve annotator agreement. 

\subsection{Annotation guidelines and annotation interface}

\begin{figure*}[h]
    \centering
    \includegraphics[width=\linewidth,page=1]{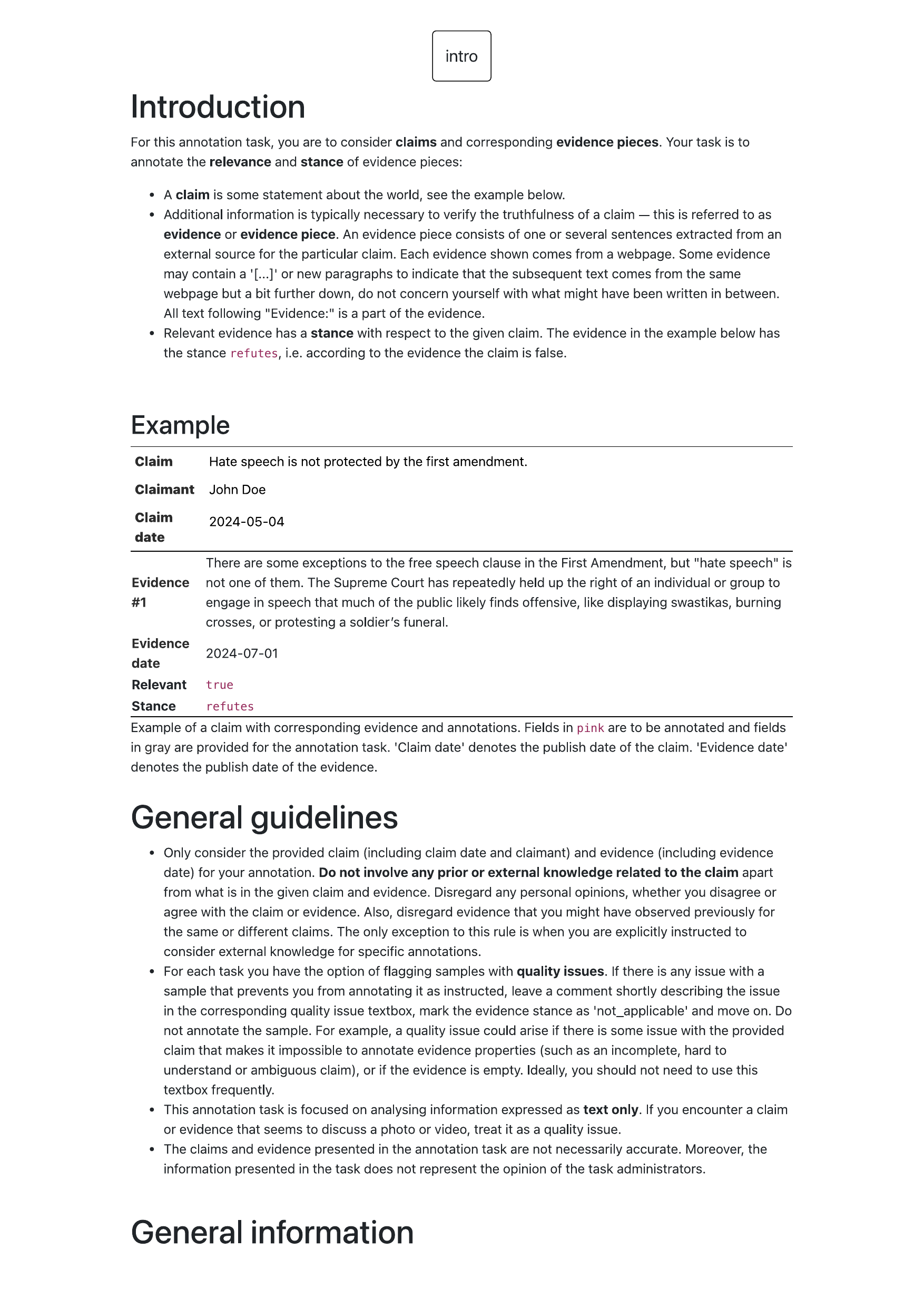}
    \caption{Page 1 of 10 depicting the annotation interface.}
    \label{fig:potato-1}
\end{figure*}

\begin{figure*}[h]
    \centering
    \includegraphics[width=\linewidth,page=2]{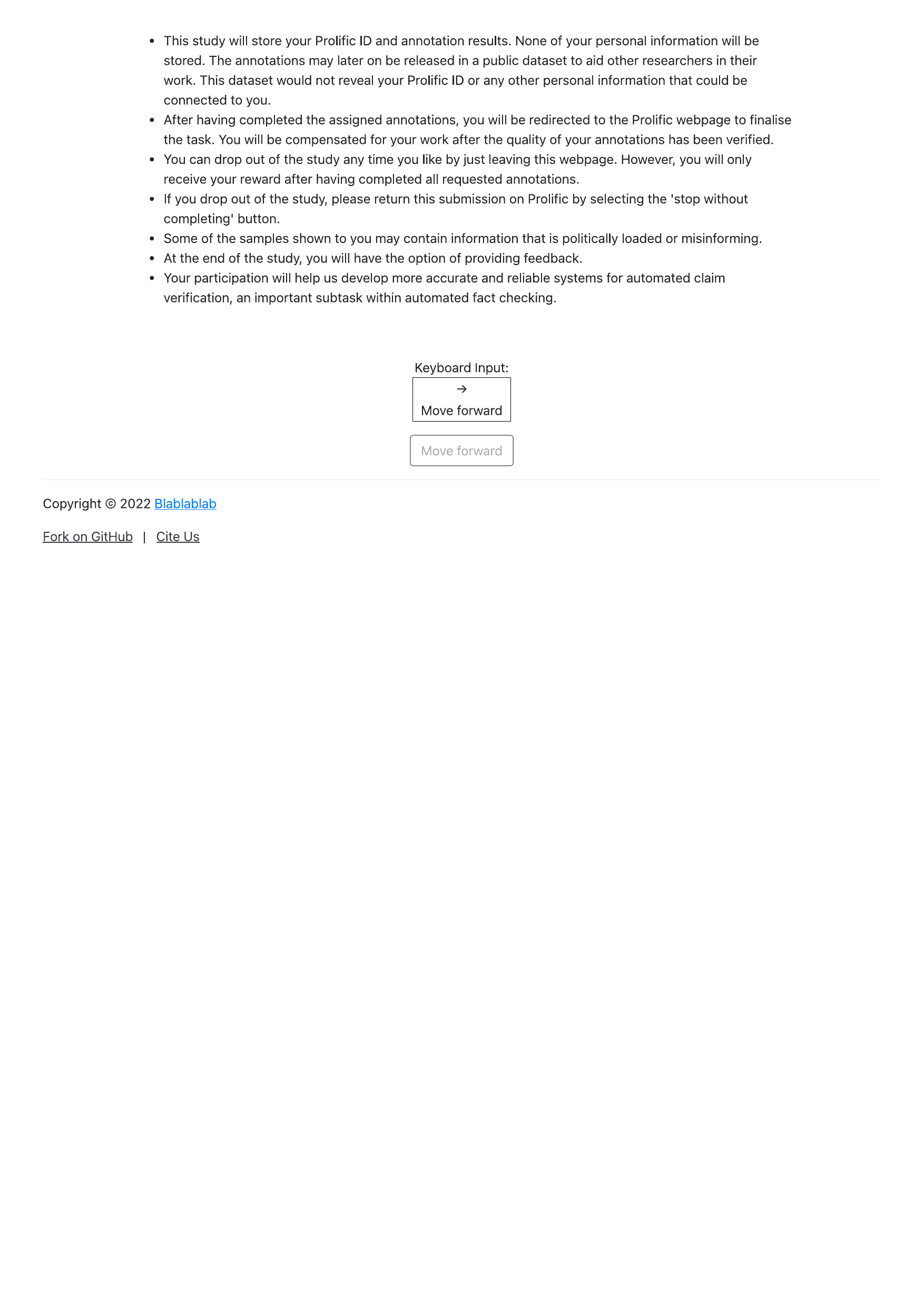}
    \caption{Page 2 of 10 depicting the annotation interface.}
    \label{fig:potato-2}
\end{figure*}

\begin{figure*}[h]
    \centering
    \includegraphics[width=\linewidth,page=3]{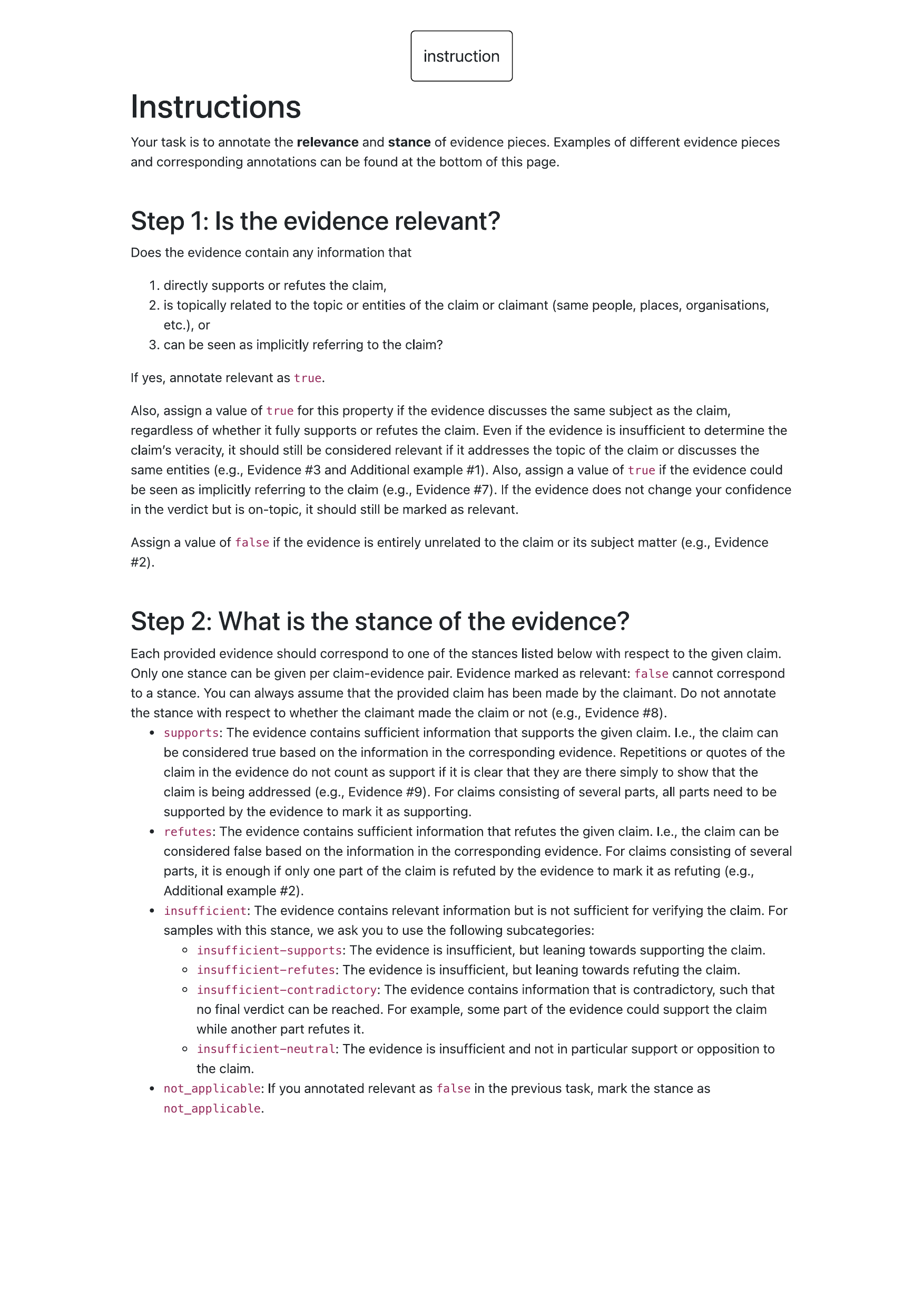}
    \caption{Page 3 of 10 depicting the annotation interface.}
    \label{fig:potato-3}
\end{figure*}

\begin{figure*}[h]
    \centering
    \includegraphics[width=\linewidth,page=4]{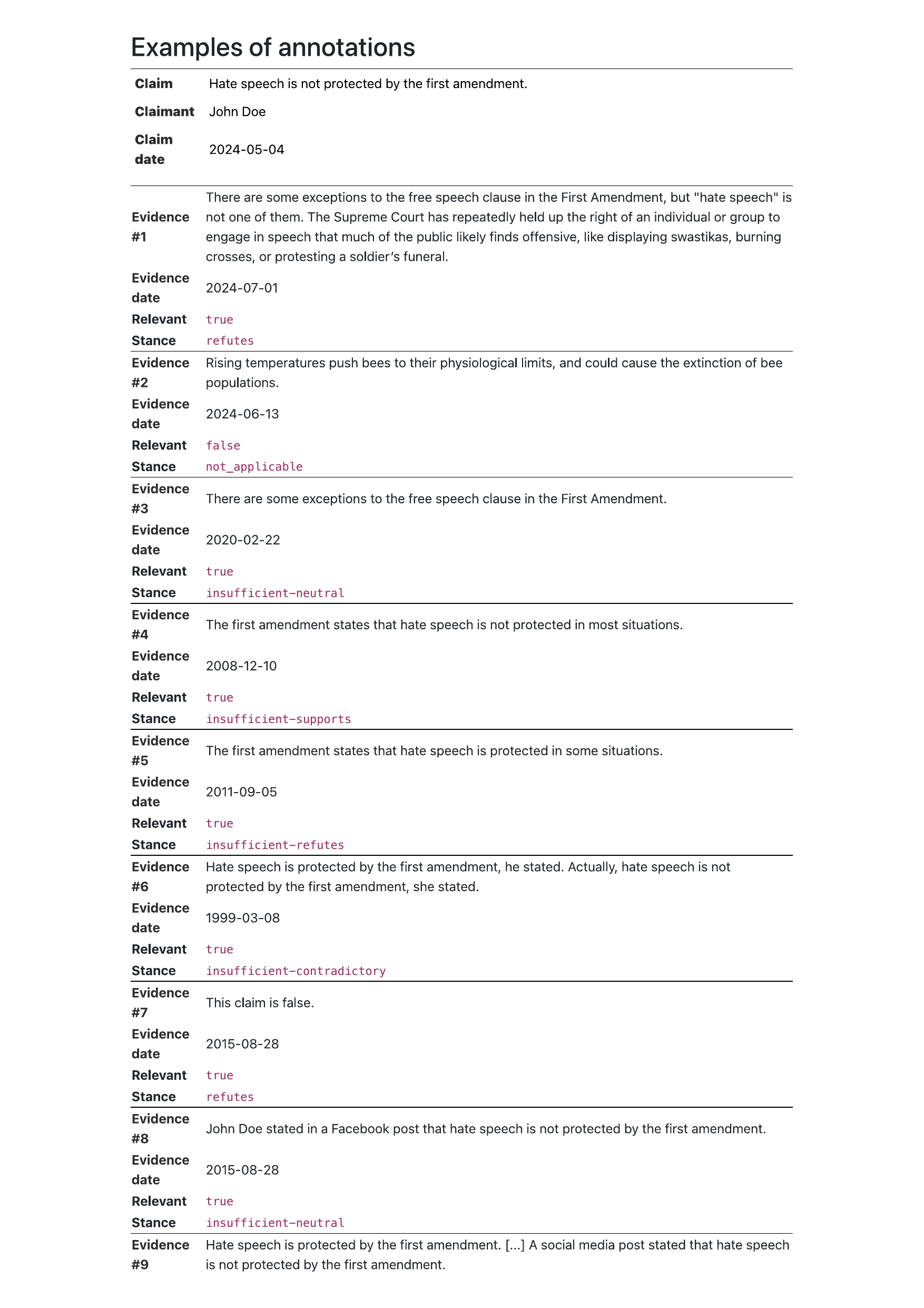}
    \caption{Page 4 of 10 depicting the annotation interface.}
    \label{fig:potato-4}
\end{figure*}

\begin{figure*}[h]
    \centering
    \includegraphics[width=\linewidth,page=5]{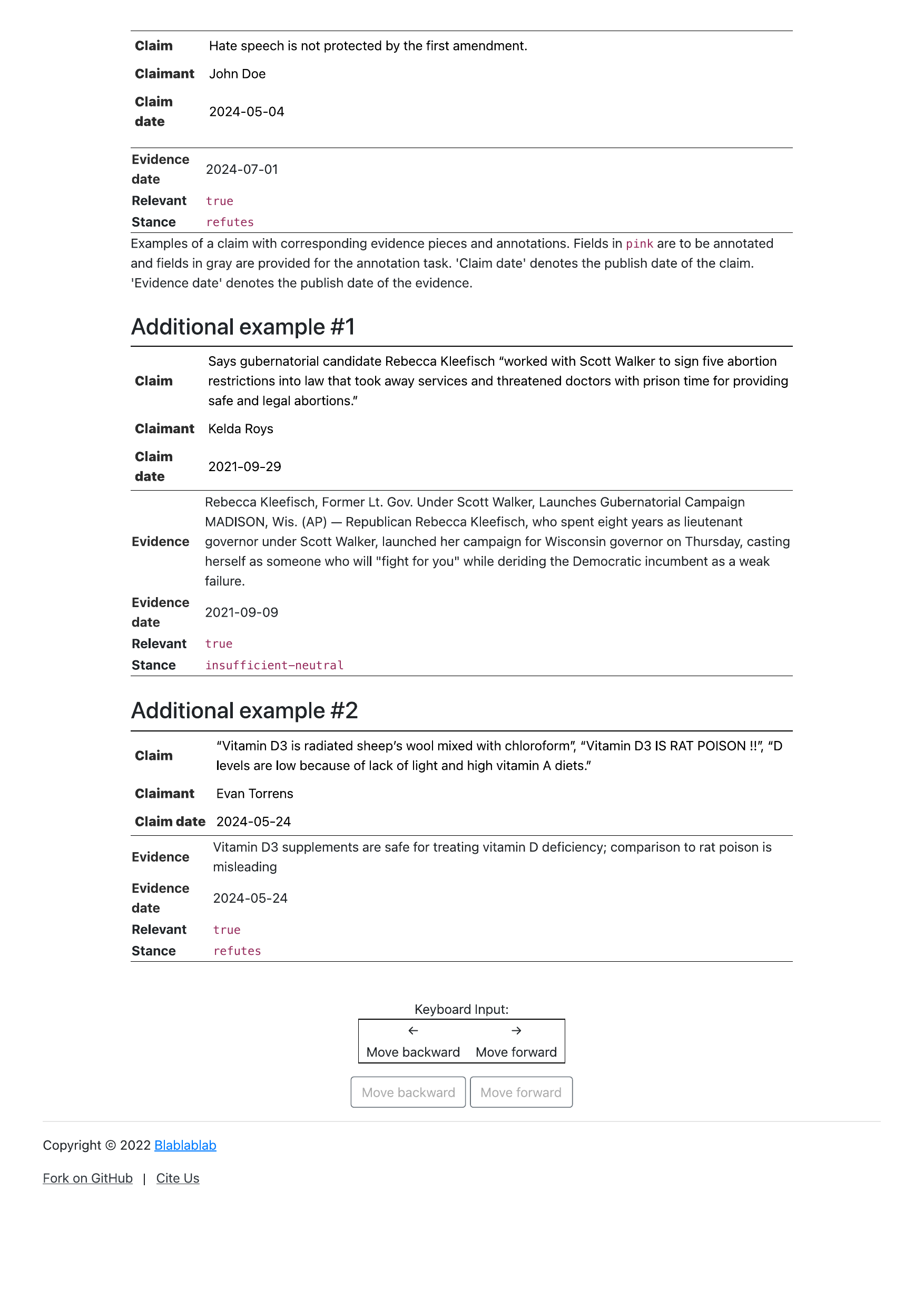}
    \caption{Page 5 of 10 depicting the annotation interface.}
    \label{fig:potato-5}
\end{figure*}

\begin{figure*}[h]
    \centering
    \includegraphics[width=\linewidth,page=6]{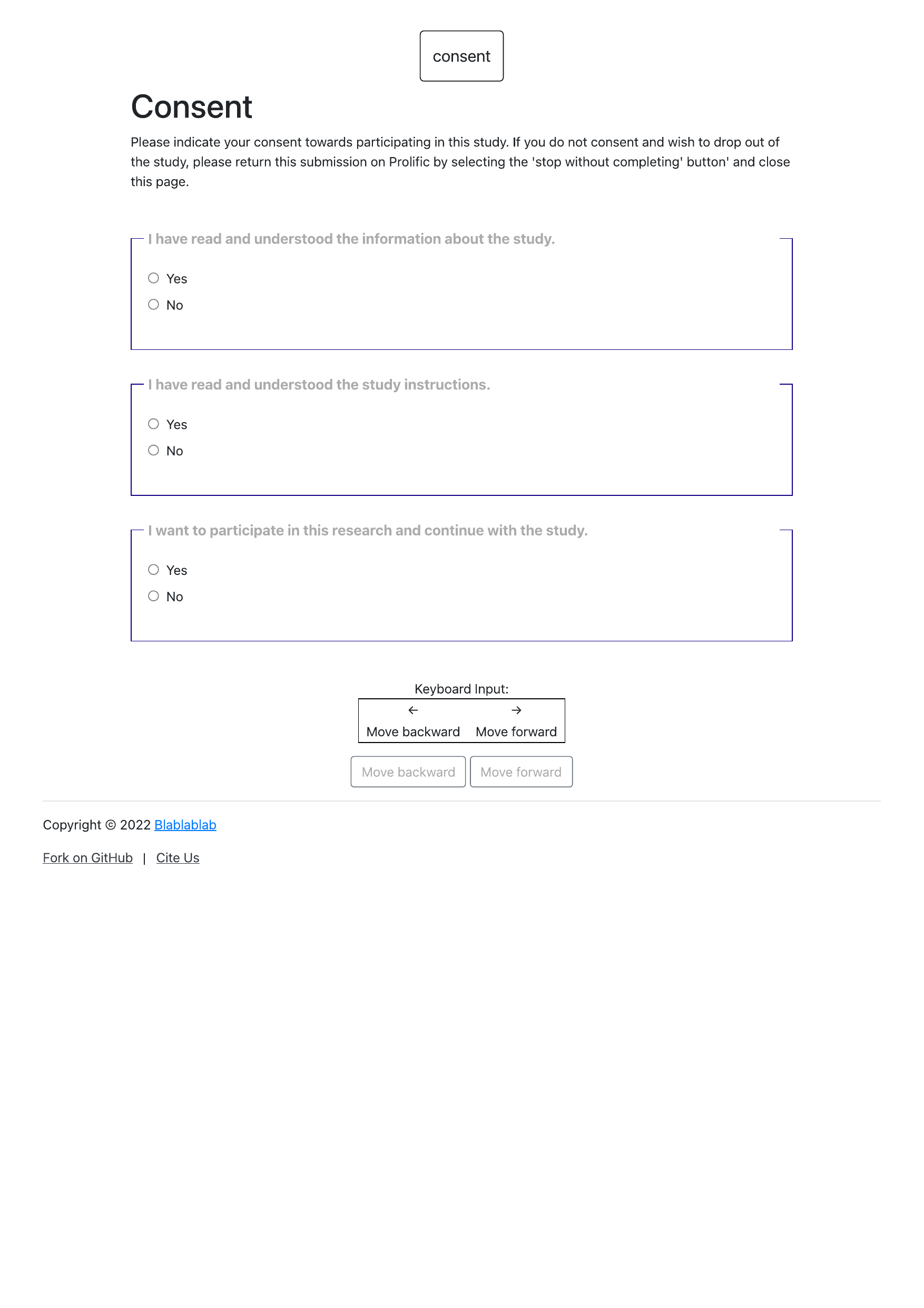}
    \caption{Page 6 of 10 depicting the annotation interface.}
    \label{fig:potato-6}
\end{figure*}

\begin{figure*}[h]
    \centering
    \includegraphics[width=\linewidth,page=7]{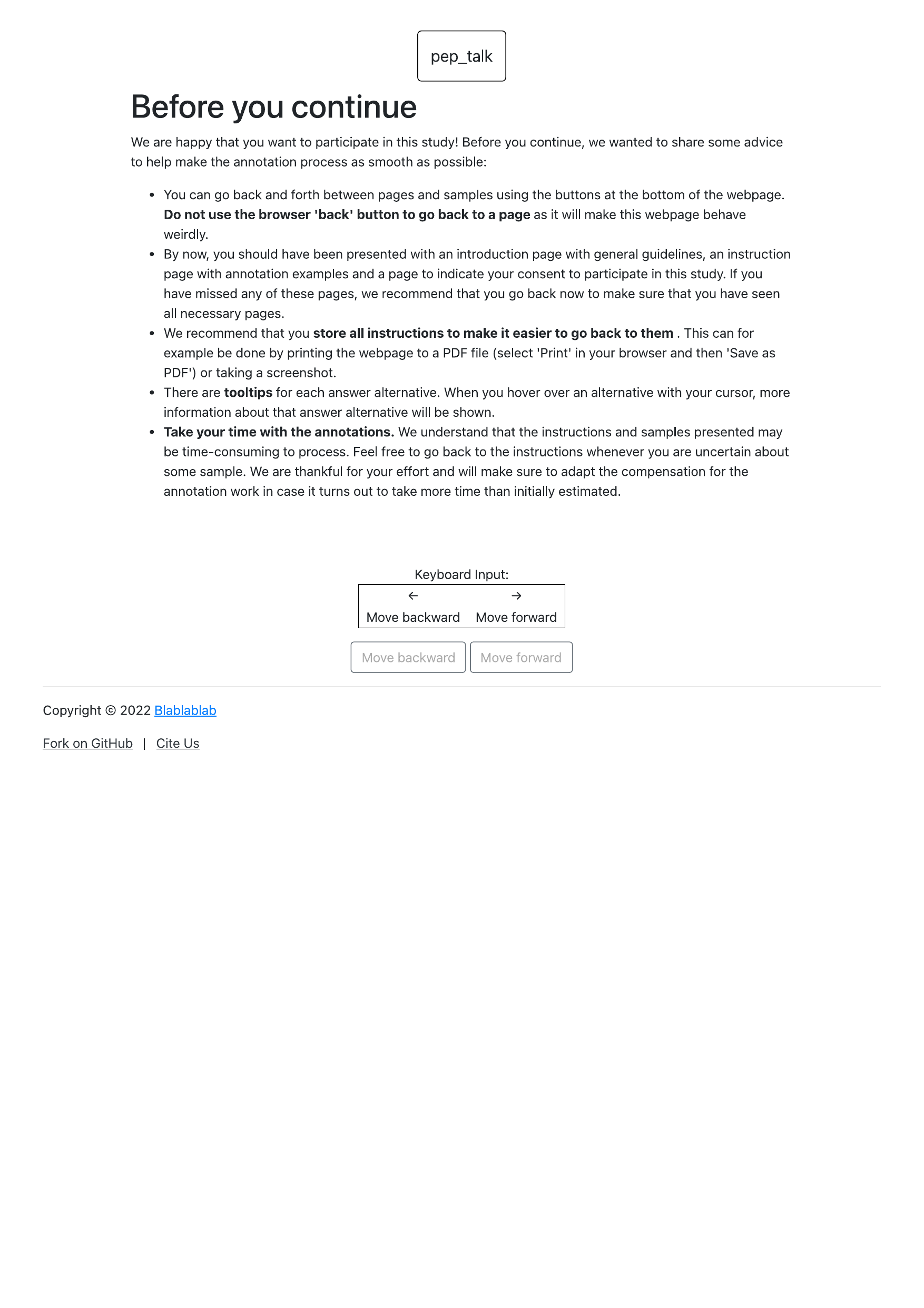}
    \caption{Page 7 of 10 depicting the annotation interface.}
    \label{fig:potato-7}
\end{figure*}

\begin{figure*}[h]
    \centering
    \includegraphics[width=\linewidth,page=8]{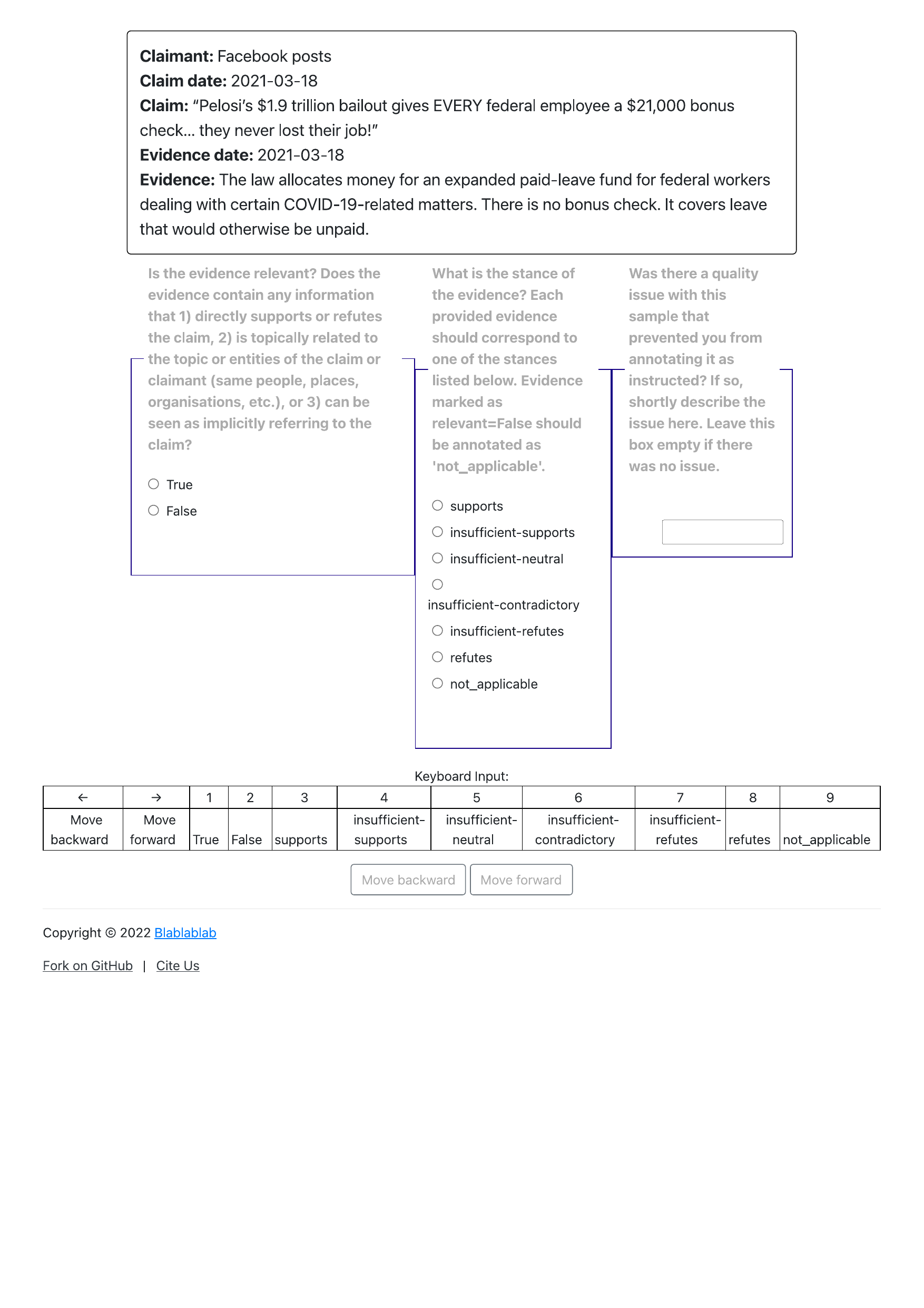}
    \caption{Page 8 of 10 depicting the annotation interface.}
    \label{fig:potato-8}
\end{figure*}

\begin{figure*}[h]
    \centering
    \includegraphics[width=\linewidth,page=9]{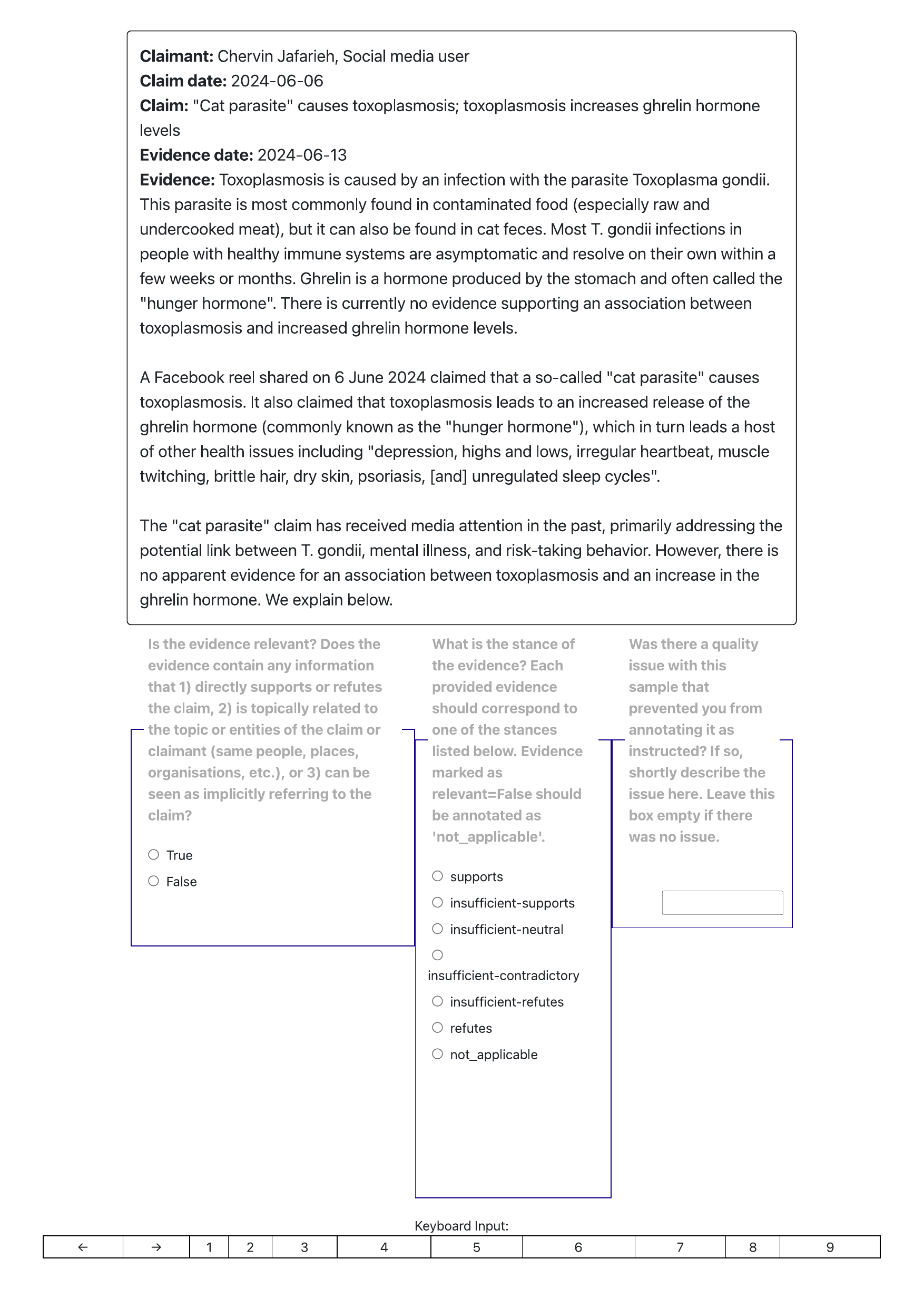}
    \caption{Page 9 of 10 depicting the annotation interface.}
    \label{fig:potato-9}
\end{figure*}

\begin{figure*}[h]
    \centering
    \includegraphics[width=\linewidth,page=10]{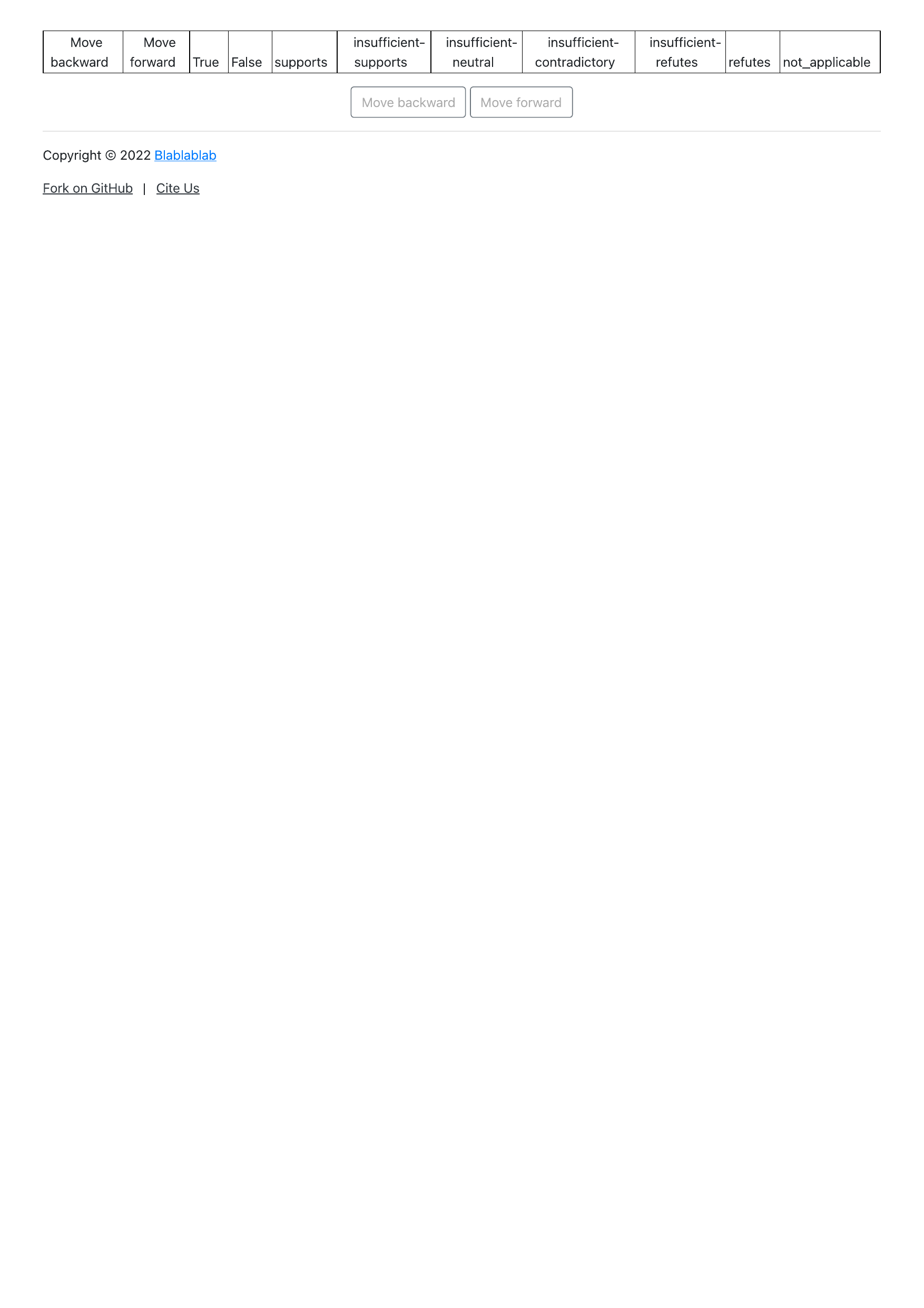}
    \caption{Page 10 of 10 depicting the annotation interface.}
    \label{fig:potato-10}
\end{figure*}

The annotation guidelines and two examples annotation pages from the annotation interface can be found in \Cref{fig:potato-1,fig:potato-2,fig:potato-3,fig:potato-4,fig:potato-5,fig:potato-6,fig:potato-7,fig:potato-8,fig:potato-9,fig:potato-10}.
\clearpage

\end{document}